\documentclass[11pt]{article}
\usepackage{fullpage}
\usepackage{authblk}

\usepackage{cuted}
\usepackage{algorithm}
\usepackage{algorithmic}
\usepackage[utf8]{inputenc}
\usepackage[english]{babel}
\usepackage{amsthm}
\theoremstyle{plain}
\usepackage[utf8]{inputenc} % allow utf-8 input
\usepackage{hyperref}       % hyperlinks

\usepackage{microtype}
\usepackage{graphicx}

\usepackage{booktabs} 
\usepackage{epstopdf}
\usepackage{tcolorbox}
\usepackage{array}
\usepackage{hyperref}
\usepackage{amsmath,amssymb}
\usepackage{tikz}
\usepackage{float}

\usepackage{comment}
\usepackage{url}

\usepackage{natbib}
\usepackage{tikz}
\usepackage{xcolor}
\usetikzlibrary{arrows}
\usepackage{hyperref}

\newcommand{\mat}{\mathbf}
\newcommand{\vect}{\mathbf}

\def\vw{\mathbf{w}}
\def\va{\mathbf{a}}

\usepackage{hyperref}

\usepackage{tcolorbox}

\usepackage[font=small,labelfont=bf]{caption}

\usepackage{array}
\newcolumntype{P}[1]{>{\centering\arraybackslash}p{#1}}

\def\vx{\mathbf{x}}
\def\vy{\mathbf{y}}
\def\vw{\mathbf{w}}
\def\vW{\mathbf{W}}

\def\v0{\mathbf{0}}
\def\va{\mathbf{a}}
\def\vu{\mathbf{u}}

\def\vv{\mathbf{v}}

\def\vH{\mathbf{H}}
\def\vE{\mathbf{E}}

\def\vu{\mathbf{u}}

\def\norm#1{\|#1\|}

\newcommand{\abs}[1]{\left|#1\right|}
\newcommand{\expect}{\mathbb{E}}

\newcommand{\indict}{\mathbb{I}}

\newcommand{\relu}[1]{\sigma\left(#1\right)}

\newtheorem{thm}{Theorem}[section]
\newtheorem{lem}{Lemma}[section]

\newtheorem{prop}{Proposition}[section]
\newtheorem{asmp}{Assumption}[section]
\newtheorem{defn}{Definition}[section]

\newtheorem{rem}{Remark}[section]

\newtheorem{condition}{Condition}[section]

\newtheorem*{lem*}{Lemma}
\newtheorem*{con*}{Condition}
\newtheorem*{prop*}{Proposition}

\usepackage{xpatch}
\makeatletter
\xpatchcmd{\@thm}{\thm@headpunct{.}}{\thm@headpunct{}}{}{}
\makeatother

\title{
Global Convergence of Adaptive Gradient Methods for An Over-parameterized Neural Network
}
\author[1]{Xiaoxia Wu}
\author[2]{Simon S. Du}
\author[1]{Rachel Ward}
\affil[1]{Department of Mathematics, The University of Texas at Austin}
\affil[2]{Machine Learning Department, Carnegie Mellon University}
\begin{document}

\maketitle

\begin{abstract}
\label{sec:abs}
%\simon{
%Selling points:
%\begin{itemize}
%\item Improved learning rate (maybe we can say something on the optimality).
%\item First global convergence analysis of the adaptive gradient methods for neural networks.
%Emphasize on non-smoothness, non-convexity, etc.
%\item An interleaving analysis of over-parameterization and adaptivity.
%\end{itemize}		
%%	
%General comments:
%\begin{itemize}
%	\item Except $\alpha,\eta,b_0$, don't use other letters like $L$.
%\end{itemize}	
%}

Adaptive gradient methods like AdaGrad are widely used in optimizing neural networks.  Yet, existing convergence guarantees for adaptive gradient methods require either convexity or smoothness, and, in the smooth setting, only guarantee convergence to a stationary point.   We propose an adaptive gradient method and show that for two-layer over-parameterized neural networks -- if the width is sufficiently large (polynomially) -- then the proposed method converges \emph{to the global minimum} in polynomial time, and convergence is robust, \emph{ without the need to fine-tune hyper-parameters such as the step-size schedule and  with the level of over-parametrization independent of the training error}.  Our analysis indicates in particular that over-parametrization is crucial for the harnessing the full potential of adaptive gradient methods in the setting of neural networks.   
%Notably, our method is robust to hy

%Over-parametrization nueral networks with stochastic gradient descent have gained widespread interests for their ability to converge globally and achieve low generalization error. Recent theoretical guarantees for this surprising phenomenon, yet, are restrictive in small constant learning rate that polynomially depends on the size of training data, far from what is being used in practice.  In this paper, based on the framework of two-layer fully connected ReLU activated neural networks, we assure that the learning rate can be optimally improved to depend linearly on $\frac{1}{\|\vH\|}$, the maximum eigenvalue of Gram matrix induced by this network.  

%Due to the high computational cost of $\|\vH\|$ for large-scale data,  adaptive  gradient methods are proposed to auto-tune the learning rate on-the-fly according to the gradients received along the way.  Yet,  to our best knowledge, there is no theoretical guarantees to date for non-smoothness and  non-convex  objective function in  neural networks. We give first global convergence guarantee of the adaptive gradient methods and  provide interleaving analysis of over-parameterization and adaptivity for the two-layer fully connected ReLU activated neural networks

%\simon{TO DOs:
%	\begin{itemize}
%		\item Check references.
%		\item Check $\|\cdot\|$ is used properly.
%		\item Check $H_1,H_2$ are replaced by Assumption~\ref{asmp:norm1} and~\ref{asmp:lambda_0}.
%		\item Replace $I$ by $\mat{I}$
%	\end{itemize}
%	}

\end{abstract}

\section{Introduction}
\label{sec:intro}
Gradient-based methods are widely used in optimizing neural networks.
One crucial component in gradient methods is the learning rate (a.k.a. step size) hyper-parameter, which determines the convergence speed of the optimization procedure. A large learning rate can speed up the convergence but if it is larger than a threshold, the optimization algorithm cannot converge.  This is by now well-understood for convex problems; excellent works on this topic include \cite{nash1991numerical}, \cite{bertsekas1999nonlinear}, \cite{nesterov2005smooth}, \cite{haykin2005cognitive}, \cite{bubeck2015convex}, and the recent review for large-scale stochastic optimization to \cite{bottou2018optimization}. However, there is still limited work on the convergence analysis for nonsmooth and nonconvex problems, which includes over-parameterized neural networks.

Recently, a series of breakthrough papers showed that (stochastic) gradient descent can provably converge to the global minima for over-parameterized neural networks~\citep{du2018gradient,du2018deep,li2018learning,allen2018convergence,zou2018stochastic}.
However, these papers all require the step size to be sufficiently small to guarantee the global convergence.
In practice, these optimization algorithms can use a much larger learning rate while still converging to the global minimum.
%However, practitioners want to use large learning rate to speed up the convergence.
%While existing papers provide global convergence rate result, the learning rate is often small, like $O(1/n^2)$ for average loss, which is far from what is being used in practice. 
This leads to the following question: \begin{center}
\emph{What is the optimal learning rate in optimizing neural networks?}
\end{center}

While finding the optimal step size is important theoretically for identifying the optimal convergence rate, the optimal learning rate often depends on certain unknown parameters of the problem. For example, for a convex and $L$-smooth objective function, the optimal learning rate is $O(1/L)$ where $L$ is often unknown to practitioners. To solve this problem, adaptive methods~\citep{duchi2011adaptive,mcmahan2010adaptive} are proposed so that they can change the learning rate on-the-fly according to gradient information received along the way. 
Though these methods often introduce additional hyper-parameters, compared to gradient descent methods with well-tuned stepsize, the adaptive methods are often robust to their hyper-parameters in the sense that these methods can still converge modulo (slightly) slower convergence rate.
For this reason, adaptive gradient methods are widely used by practitioners in neural network optimization.

On the other hand, the theoretical investigation in adaptive methods in optimizing neural networks is limited. Existing analyses only deal with general (non)-convex and smooth functions, and thus, only concern convergence to first-order stationary points.  However, a neural network is \emph{neither smooth nor convex}.  And yet, adaptive gradient methods are widely used in this setting as they converge without requiring a fine-tuned learning rate schedule.   This leads to the following question:
\begin{center}
\emph{What is the convergence rate of adaptive gradient methods in over-parameterized networks?  } 
\end{center}
In this paper, we make progress on these two problems for the two-layer over-parameterized ReLU-activated neural networks setting.

\paragraph{Our Results}
\begin{itemize}
	\item First, we show the learning rate of gradient descent can be improved to $O(1/\norm{\mat{H}^{\infty}})$ where $\mat{H}^{\infty}$ is a Gram matrix that only depends on the data. 
	Note that this upper bound is independent of the number of parameters.
	As a result, using this stepsize, we show gradient descent enjoys a faster convergence rate.
	This choice of stepsize directly leads to an improved convergence rate compared to \cite{du2018gradient}.
	\item We develop an adaptive gradient method, which can be viewed as a variant of the ``norm" version of AdaGrad.
	We prove this adaptive gradient method converges to the global minimum in polynomial time and does so robustly, in the sense that for any choice of hyper-parameters used in this method, our method is guaranteed to converge to the global minimum in polynomial time.
	The choice of hyper-parameters only affect the rate but not the convergence.
	To our knowledge, this is the first polynomial time global convergence result for an adaptive gradient method in the non-convex setting.
\end{itemize}

\paragraph{Challenges and Our Techniques}
To verify the improved learning rate of gradient descent, we use a more subtle  analysis of the dynamics of predictions considered in \cite{du2018gradient}.
Our analysis shows that the dynamics are close to a linear one.
This observation allows us to choose the improved learning rate.

For the adaptive method, there are two big challenges.
First, because the learning rate (induced by the hyper-parameters and the dynamics) is changing at every iteration, we need to lower and upper bound the learning rate.
The lower bound is required to guarantee the algorithm will converge in polynomial time and the upper bound is required to guarantee the algorithm will not diverge.
The second challenge is that if at the beginning the learning rate is too large, the loss may increase at the beginning.
The proof of \cite{du2018gradient} for gradient descent with well-tuned stepsize highly depends on the fact that the loss is decreasing geometrically at each iteration, so that proof cannot be adapted to our setting.

In this paper, we use induction with a carefully constructed hypothesis which implies both the upper and the lower bounds of the learning rate.
%\simon{@shirley, }
Furthermore, utilizing the particular property induced by our proposed adaptive algorithm, the learning rate learns from feedback from previous iterations and thus perseveres the distance of the updated weight matrix and its initialization (Lemma \ref{lem: small_w_for_allbsq1_sub}) while does not vanishes to zero (Lemma \ref{lem: small_w_for_allbsq_sub}).
This property, together with the effect of over-parameterization, we show that the loss may only increase by a bounded amount and then decreases to zero eventually.
Resolving these issues, we are able to prove the first global convergence result for an adaptive gradient method in optimizing neural networks.

%\simon{@shirley, can you write a draft? I}

\subsection{Related Work}
\paragraph{Global Convergence of Neural Networks}
\label{sec:rel}
Recently, a series of papers showed that gradient based methods can provably reduce the training error to $0$ for over-parameterized neural networks~\citep{du2018gradient,du2018deep,li2018learning,allen2018convergence,zou2018stochastic}
. In this paper we study the same setting considered in \cite{du2018gradient} which showed that for learning rate $\eta = O(\lambda_{\min}(\mat{H}^{\infty})/n^2)$, gradient descent finds an $\varepsilon$-suboptimal global minimum in $O\left(\frac{1}{\eta\lambda_{\min}(\mat{H}^{\infty})}\log(\frac{1}{\epsilon})\right)$ iterations for the two-layer over-parameterized ReLU-activated neural network.  As a by-product of the analysis in this paper, we show that the learning rate can be improved to $\eta = O(1/\norm{\mat{H}^{\infty}})$ which results in faster convergence.
%Moreover, we develop an adaptive gradient method for this setting and prove the polynomial time convergence guarantee.
We believe that the proof techniques developed in this paper can be extended to deep neural networks, following the recent works ~\citep{du2018deep,allen2018convergence,zou2018stochastic}.

\paragraph{Adaptive Gradient Methods} 
%\simon{@Shirley, can you write a complete draft? I can polish it after your pass.}
Adaptive Gradient (AdaGrad) Methods, first introduced independently by \citet{duchi2011adaptive} and \citet{mcmahan2010adaptive}, are now widely used in practice for online learning due in part to their robustness to the choice of stepsize. The first convergence guarantees, proved in \citet{duchi2011adaptive}, were for the setting of online convex optimization  where the loss function may change from iteration to iteration.  Later convergence results for the variants of AdaGrad were proved in \citet{levy2017online} and \citet{mukkamala2017variants} for offline convex and strongly convex settings.   In the general non-convex and smooth setting, \citet{ward2018adagrad} and  \citet{li2018convergence} prove that the same ``norm" version of AdaGrad converges to a stationary point at rate $O\left({1}/{\varepsilon^2}\right)$ for stochastic gradient descent and at rate $O\left( {1}/{\varepsilon}\right)$ for batch gradient descent.

Many modifications to AdaGrad have been proposed, namely,  RMSprop \citep{hinton2012neural}, AdaDelta \citep{zeiler2012adadelta}, Adam \citep{kingma2014adam}, AdaFTRL\citep{orabona2015scale}, SGD-BB\citep{tan2016barzilai}, AdaBatch \citep{defossez2017adabatch}, signSGD \citep{pmlr-v80-bernstein18a},  SC-Adagrad \citep{mukkamala2017variants,Shah2018MinimumNS}, WNGrad \citep{wu2018wngrad}, AcceleGrad \citep{OnlineLevy2018}, Yogi \citep{aheer2018adaptive1}, Padam \citep{chen2018closing}, to name a few.  More recently, acccelerated adaptive gradient methods have also been proved to converge to stationary points~\citep{barakat2018convergence, chen2018on, ma2018adaptive,  zhou2018convergence,zou2018sufficient}. 
%However, these accelerated methods are not the focus in this paper.

Our work is inspired by the analysis of \cite{ward2018adagrad} and \cite{wu2018wngrad} which quantifies the auto-tuning property in the learning rate in AdaGrad.  
We propose a new adaptive algorithm for the stepsize in the setting of over-parameterized neural networks and show global polynomial convergence guarantee.  
%We emphasize that in our analysis, the \emph{level of overparameterization is independent of the training error} and, unlike previous convergence results in this setting using fixed step-size, the convergence results are \emph{robust to unknown parameters, such as the eigenvalues of the data Gram matrix}. 

\section{Problem Setup}
\label{sec:pre}
%\subsection{Setup for analysis}
%\simon{The analyses in this section already appeared in \cite{du2018gradient}. Please move them to appendix.}
\paragraph{Notations}
Throughout, $\| \cdot \|$ denotes the Euclidean norm if it applies to a vector  and the maximum eigenvalue if it applies to a matrix. 
%$\mat{I}$ is identity matrix.
%\simon{Is $\norm{\cdot}$ for spectral norm standard? Or we should use $\|\cdot\|_{2}$?}
%\shirley{we use this for our adaptive gradient sharp convergence paper..}
We use $N(\vect{0},\mat{I})$ to denote a standard Gaussian distribution where $\mat{I}$ denotes the identity matrix and $U(S)$ to denote the uniform distribution over a set $S$.
 We use the notation $[n] := \{ 0,1,2, \dots, n\}$.  
% We use $O(\cdot)$ and $\Omega\left(\cdot\right)$ to denote the usual Big-O and Big
% $\widetilde{O}$ or $\widetilde{\Omega}$ means   ${O}$ or $\Omega$  with the hidden logarithmic order.
%% \simon{@Shirley: where do we use $\widetilde{O}$ and $\widetilde{\Omega}$? If not let's stick to $O$ and $\Omega$}
% \shirley{we use $\widetilde{O}$ in Theorem 4.1 Case (1) where we only use $\log(1/\varepsilon) $  instatead should be $\log(n/\varepsilon) $ as well as case (2)... with more log terms }
% \simon{let's explain $\widetilde{O}$ and $\widetilde{\Omega}$ in the footnote of the theorem.}

 \paragraph{Problem Setup}
 In this paper we consider the same setup as \cite{du2018gradient}.
We are given $n$ data points, $\{\vect{x}_i,y_i\}_{i=1}^n$.
Following \cite{du2018gradient}, to simplify the analysis, we make the following assumption on the training data.
\begin{asmp}
	\label{asmp:norm1}
For $i \in [n]$, $\norm{\vect{x}_i} = 1$ and $\abs{y_i} = O(1)$.
\end{asmp}
The assumption on the input is only for the ease of presentation and analysis.
See discussions in \cite{du2018gradient}.
The second assumption on labels is satisfied in most real world datasets.
 
We predict labels using a  two-layer neural network of the following form
 \begin{align}
     f(\vW,\va,\vx) = \frac{1}{\sqrt{m}} \sum_{r=1}^{m}a_r
     \sigma(\langle{\vw_r, \vx\rangle}) 
     \label{eq:averagem}
 \end{align}
where $\vx\in \mathbb{R}^d$ is the input, for $r \in [m]$, $\vw_r \in \mathbb{R}^d$ the weight vector  of the first layer and $a_r \in \mathbb{R}$ is the output weight  and $\sigma(\cdot)$ is ReLU activation function. 
For $r \in [m]$, we initialize the first layer vector $\vect{w}_r(0) \sim N(\vect{0},\mat{I})$ and output weight $a_r \sim U(\left\{-1,+1\right\})$.
We fix the second layer and train the first layer with the quadratic loss
 \begin{align}
\label{eq:loss_ave}
    L(\vW) =\sum_{i=1}^{n}
     \frac{1}{2}(f(\vW,\va,\vx_i)-y_i)^2.
 \end{align}
 We will use iterative gradient-based algorithms to train $\mat{W}$.
 The gradient of each weight vector has the following form:
\begin{align}
   \label{eq:grad_ave}
   \frac{\partial L(\vW)}{\partial \vw_r} = \frac{a_r}{\sqrt{m}}\sum_{i=1}^{n}
   (f(\vW,\va,\vx_i)-y_i)\vx_i\mathbb{I}_{\{\vw_r^T\vx_i\geq0\}}
\end{align} 
  We use $\mat{W}(k)$ to denote the parameters at the $k$-th iteration.
 
 The training algorithm will be specified in Section~\ref{sec:gd} and~\ref{sec:adaloss}.
 Define  $u_i =f(\vW,\va,\vx_i)$, the prediction of the $i$-th example and $\vect{u} =\left(u_1,\ldots,u_n\right)^\top \in \mathbb{R}^n$.
 We also let $\vect{y} = \left(y_1,\ldots,y_n\right)^\top \in \mathbb{R}^{n}$.
  Then we can write the loss function as 
 \begin{align*}
    L(\vW) = \frac{1}{2}\|\vu-\vy\|^2.
 \end{align*}
 In this paper, we will study the dynamics of $\vect{u}(k)$.
 Here we use $k$ for indexing because $\vect{u}(k)$ is induced by $\mat{W}(k)$.
% While instead of understanding $\|\vW({k}) - \vW^{*}\|_2$ where $\vW^{*}$ is  optimal solution, we look at the $\|\vu{(k)} - \vy\|$ so that to explain why loss function have zero training error. 
According to \cite{du2018gradient}, the matrix below determines the convergence rate of the randomly initialized gradient descent.
\begin{defn}
\label{def:H}
The matrix  $\mat{H}^\infty \in \mathbb{R}^{n \times n}$ is defined as follows.
For $(i,j) \in [n] \times [n]$.
\begin{align}
\mat{H}_{ij}^\infty = \expect_{\vect{w} \sim N(\vect{0},\mat{I})}\left[\vect{x}_i^\top \vect{x}_j\indict\left\{\vect{w}^\top \vect{x}_i \ge 0, \vect{w}^\top \vect{x}_j \ge 0\right\}\right] = \vect{x}_i^\top \vect{x}_j\frac{\pi - \arccos \left(\vect{x}_i^\top \vect{x}_j\right)}{2\pi}\label{eqn:H_inf}
\end{align}
\end{defn}
This matrix represents the kernel matrix induced by Gaussian initialization and ReLU activation function.
We make the following assumption on $\mat{H}^{\infty}$.
\begin{asmp}\label{asmp:lambda_0}
	The matrix  $\mat{H}^\infty \in \mathbb{R}^{n \times n}$ in Definition \ref{def:H} satisfies $\lambda_{\min}(\mat{H}^{\infty}) \triangleq \lambda_ 0 > 0$.
\end{asmp}
\citet{du2018gradient} showed that this condition holds as long as the training data is not degenerate.
We also define the following empirical version of this Gram matrix, which will be used in our analysis.
For $(i,j) \in [n] \times [n]$:
\begin{align}
\mat{H}_{ij} = \frac{1}{m}\sum_{r=1}^{m}\vect{x}_i^\top \vect{x}_j \indict{\left\{\vect{w}_r^\top \vect{x}_i \ge 0, \vect{w}_r^\top \vect{x}_j \ge  0\right\}}. \label{eqn:H}
 \end{align}
%\paragraph {The assumptions} 
%We make several assumptions that are general and close to practice. 
%\begin{itemize}
%\item [\textbf{(H1)}] The training dataset $\{\vx_i, y_i\}$ are normalized, i.e., $i \in [n]$, $\norm{\vect{x}_i}_2 = 1$ and $\abs{y_i} \le C_y$ for some constant $C_y.$ 
%\end{itemize}
% As  the gradient of $W$, $ \frac{\partial L(\vW(k))}{\partial \vW }$, is matrix in $\mathbb{R}^{d\times m}$, we  express the gradient of each neuron  $r\in [m]$ at iteration $k$,

\section{Warm up: Improved Learning Rate for Gradient Descent}
\label{sec:gd}
 Before presenting our adaptive method, we first revisit the gradient descent algorithm.
 At each iteration $k = 0,1,\ldots$,  we update the weight matrix according to \begin{align}
\vW{(k+1)} = \vW{(k)} - \eta \frac{\partial L(\vW{(k)})}{\partial \vW }\label{eqn:gd}
 \end{align}
 where $\eta > 0$ is the learning rate.
 \cite{du2018gradient} showed if $\eta = O(\lambda_0/n^2)$, then gradient descent achieves $0$ training loss at a linear rate.
 We improve the upper bound of learning rate used in \cite{du2018gradient}.
 This improved analysis also gives tighter bound for the adaptive method we will discuss in the next section.
 Our main result for gradient descent is the following theorem.
% \simon{We may move assumptions and definitions out of the theorem to make the statement cleaner}
\begin{thm}
[Convergence Rate of Gradient Descent with Improved Learning Rate]
\label{thm:main_gd}
 Under Assumption~\ref{asmp:norm1} and~\ref{asmp:lambda_0}, if the number of hidden nodes $m =
\Omega\left(\frac{n^6}{\lambda_0^4 \delta^{3}}\right)
$ and  we set the stepsize to be 
\[ \eta = \Theta \left ( \frac{1}{ \norm{\vH^{\infty}}} \right),\]
 then with probability at least $1-\delta$
 over the random initialization, after\footnote{
 			$\widetilde{O}$ and $\widetilde{\Omega}$ hide $\log(n),\log(1/\lambda_0), \log(1/\delta)$ terms.
 		}
\begin{align*}
T =\widetilde{O} \left( \frac{ \norm{ \vH^{\infty}}}{   \lambda_0 } \log\left( \frac{ 1 }{ \varepsilon} \right) \right)
\end{align*} 
iterations, we have 
$L(\mat{W}(T)) \le \varepsilon$. 
\end{thm}

Comparing with \cite{du2018gradient}, we improve the maximum allowable learning rate from $O(\lambda_0/n^2)$ to $O(1/\norm{\mat{H}^{\infty}})$.
Note since $\norm{\mat{H}^{\infty}} \le n$,  Theorem \ref{thm:main_gd} gives an $O(\lambda_0/n)$ improvement.
The improved learning also gives a tighter iteration complexity bound $O \left( \frac{ \norm{ \vH^{\infty}}}{   \lambda_0 } \log\left( \frac{ n }{ \varepsilon \delta }\right) \right)$ comparing to the $O \left( \frac{n^2} { \lambda_0^2} \log\left( \frac{ n }{ \varepsilon \delta }\right) \right)$ bound in \cite{du2018gradient}. 
Empirically, we found that if the data matrix is approximately orthogonal, then $\norm{\mat{H}^{\infty}} =O\left(1\right)$ (see Figure \ref{fig:data} in Appendix \ref{sec:basic}). 
%\simon{cross ref some plots}
Therefore, in certain scenarios, the iteration complexity of gradient descent is independent of  $n$.

Note even though gradient descent gives fast convergence, one needs to set the learning rate $\eta$ appropriately to achieve the fast convergence rate.
In practice, $\norm{\mat{H}^{\infty}}$ is unknown to users so it would be better if the learning rate can be automatically adjusted.
We address this problem in the next section.

\paragraph{Proof Sketch of Theorem~\ref{thm:main_gd}}
Our main observation is the following recursion formula.
\begin{align*}
\norm{\vect{y}-\vect{u}(k+1)}^2 
\approx&\norm{\vect{y}-\vect{u}(k)}^2  -   2\eta\left(\vect{y}-\vect{u}(k)\right)^\top\left(\mat{I}- \eta\mat{H}^\infty\right) \mat{H}^\infty \left(\vect{y}-\vect{u}(k)\right) \\
\le & \norm{\vect{y}-\vect{u}(k)}^2  -  2\eta \lambda_0\left(1-\eta \norm{\mat{H}^{\infty}}\right)\norm{\vect{y}-\vect{u}(k)}^2 \\
\le & \left(1-\eta \lambda_0\right) \norm{\vect{y}-\vect{u}(k)}^2 .
\end{align*}
The first approximation we used over-parameterization ($m$ is large enough) for which the width $m$ becomes larger the approximation becomes more accurate.
In Section~\ref{sec:proof_gd}, we will give precise perturbation analysis.
The first inequality we used the fact that $\eta = O\left(1/\norm{\mat{H}^\infty}\right)$ and the two symmetric matrices $\left(\mat{I}- \eta\mat{H}^\infty\right)$ and $ \mat{H}^\infty $ share same eigenvectors.
The second inequality we used $\eta = O\left(1/\norm{\mat{H}^\infty}\right)$ again.
Note this recursion formula shows the loss converges to $0$ at a linear rate and if we plug in $\eta = \Theta\left(1/\norm{\mat{H}^\infty}\right)$ we prove theorem.
The details are in Section~\ref{sec:proof_gd}.

\section{An Adaptive Method for Over-parameterized Neural Networks}
\label{sec:adaloss}
%Algorithm overview
In this section we present our new adaptive gradient algorithm for optimizing over-parameterized neural networks.
At the high level, we use the same paradigm as existing adaptive methods   \citep{duchi2011adaptive}.
There are three positive hyper-parameters, $b_0, \eta, \alpha$ in    the algorithm. $\eta$ is to ensure the homogeneity and that the units match.
$b_0$ is the initialization of a monotonically increasing sequence $\{b_k\}_{k=1}^{\infty}$ such that  $b_k$  is updated at $k$-th iteration. To control the rate of this update, we use the parameter $\alpha$. Note $\alpha$ is not the learning rate to update the parameter $\mat{W}$.
At $k$-th iteration, we first use $\alpha$ and the information received to obtain $b_{k+1}$, then use $\eta/b_{k+1}$ to update the parameters.
Here $\eta/b_{k+1}$ is the effective learning rate at the $k$-th iteration.

%motivation
In practice, we would like an adaptive method that is robust to the choices of hyper-parameters.
That is, we want this method guaranteed to converge in polynomial time for any choice of hyper-parameters.\footnote{
	The convergence rate will, of course, depend on the choices of the hyper-parameters. The convergence of the ideal adaptive algorithm only depends polynomially on the these hyper-parameters.
}
The key challenge for the adaptive method is how to design an appropriate update rule for $\{b_k\}$ to achieve the goal.
Our algorithm uses the following update rule:
\begin{align}
b_{k+1}^2 \leftarrow b_k^2 +  \alpha^2 \sqrt{n}\|\vy-\vu(k) \|. \label{eqn:bk_update}
\end{align}
Here one can just view $\alpha$ and $n$ together as one constant.  
 Using $\alpha^2$ is for matching the scale of $\eta$ and using $\sqrt{n}$ is for the ease of comparison with other adaptive gradient methods that we  further discuss in Section \ref{sec:diss}. 
 The key for this update is  $\|\vy-\vu(k) \|$  instead of its square.  Note this is sharp contrast to~\cite{duchi2011adaptive} where  the scheme to update the effective learning rate can be equivalently written as $\|\vy- \vu(k)\|^2$. The main reason is that  our  convergence analysis requires analyzing both over-parameterization and  the dynamics of  the adaptive stepsize at the same time.
%Technically, if we use $\|\vy-\vu(k) \|^2$, we can only invoke Proposition~\ref{prop:adasq} that enforces $m$ depend on the target training error $\varepsilon$ which is not desirable.
%On the other hand, if we use $\|\vy-\vu(k) \|$, we can invoke Lemma~\ref{lem
See Section~\ref{sec:diss} for more discussions.
We list pseudo codes in Algorithm~\ref{alg:adagradloss}.
\begin{algorithm}[t]
	\begin{algorithmic}
	\caption{\textbf{Ada}ptive \textbf{Loss} (AdaLoss)}
		\label{alg:adagradloss}
		\STATE {\bfseries Input:} Tolerance $\varepsilon > 0$, initialization $\mat{W}(0),\vect{a}$, positive constants $b_0, \eta$ and $ \alpha>0$.
		\STATE Set $k=0$.
		\REPEAT
		\STATE $b_{k+1}^2 \leftarrow b_k^2 +  \alpha^2 \sqrt{n}\|\vy-\vu(k) \|$  
		\STATE $ \vW{(k+1)} = \vW{(k)} - \frac{\eta}{b_{k+1}} \frac{\partial L(\vW{(k)})}{\partial \vW }$
		\UNTIL{$\|\vy-\vu(k) \|^2/2 \leq \varepsilon$} 
	\end{algorithmic}
\end{algorithm}
%Observe that  there is $\sqrt{m}$ in fact   in the rate to accumulate $b_k^2$ because the prediction $f(\vW, \va,\vx)$ in \eqref{eq:averagem} has $1/\sqrt{m}$. This is very crucial observation  for our convergence particularly when $b_0/\eta \leq C\|\vH^{\infty}\|$.  
%  We also note that in practice,  instead using one learning rate for each node, we use  $d$ learning rates for one node, i.e., for each index $\ell=1,2,\ldots,d$
%While the convergence for this algorithm is the same as Algorithm \ref{alg:adagrad} for Case (1), we have better results for Case (2) stated Corollary \ref{cor:adaloss}.
The following theorem characterizes the convergence rate of our proposed algorithm.
\begin{thm}[Convergence Rate of AdaLoss]
	\label{thm:main_adagradloss}
Under Assumption~\ref{asmp:norm1} and~\ref{asmp:lambda_0}, if the width satisfies\begin{align*}
	m = \Omega \left(\frac{n^6}{\lambda_0^4\delta^3} +  \frac{\eta^4}{\alpha^4} \frac{ n^{4} \norm{\mat H^\infty}^4}{\lambda_0^4  \delta^2}   \right).
	\end{align*}  
	Then Algorithm~\ref{alg:adagradloss}   admits  the following convergence results.
	
	\begin{itemize}
		\item If the hyper-parameter satisfies $ \frac{b_0}{\eta} \geq C \norm{\mat H^{\infty}}$, \footnote{The notation $C$  is well-defined, please check Table \ref{table} in Appendix \ref{sec:basic}} then with probability $1-\delta$ over  the random initialization  
		$\min_{t\in [T]}\norm{\vect y- \vect u(t)}^2 \le\varepsilon$ 
			%$\norm{\vect y- \vect u(t)}^2 \le\varepsilon$ 
		after
		\begin{align*}
		T = \widetilde{O}\left( \left(\frac{b_0 }{\eta \lambda_0} + \frac{\alpha^2 n}{\eta^2\lambda_0^2 \sqrt{\delta} } \right) \log\left(\frac{1}{\varepsilon} \right)\right).
		\end{align*}
\item If the hyper-parameter satisfies $ 0<\frac{b_0}{\eta} \leq C \norm{\mat H^{\infty}}$, then with probability $1-\delta$ over  the random initialization  $\min_{t\in [T]}\norm{\vect y- \vect u(t)}^2 \le\varepsilon$   after	
\begin{align*}
T =  \widetilde{O}\left( \frac{\left(\eta C\|\vH^{\infty} \|\right)^2 -b_0^2 }{\alpha^2\sqrt{ n\varepsilon} } + \left(\frac{\alpha^2 n}{\eta^2\lambda_0^2 \sqrt{\delta}} +\left( \frac{\|\vH^\infty\|}{\lambda_0} \right)^{2}\right) \log \left(\frac{1}{\varepsilon} \right)\right).
\end{align*}
	\end{itemize}
\end{thm}
To our knowledge, this is first global convergence guarantee for the adaptive gradient method.
Now we unpack the statements of Theorem~\ref{thm:main_adagradloss}.
Our theorem applies to two cases.
In the first case, the effective learning rate at the beginning $\eta/b_0$ is smaller than the threshold $1/(C \norm{\mat H^{\infty}})$ that guarantees the global convergence of gradient descent (c.f. Theorem~\ref{thm:main_gd}).
In this case, the convergence has two terms, the first term $\frac{b_0}{\eta \lambda_0}\log\left(\frac{1}{\epsilon}\right)$ is standard gradient descent rate if we use $\eta/b_0$ as the learning rate.
Note this term is the same as Theorem~\ref{thm:main_gd} if $\eta/b_0 = \Theta(1/\norm{\mat{H}^\infty})$.
The second term 
%\simon{@Shirley, please add explaination where this term is from and refer a lemma.} 
comes from the upper bound of  $b_T$ in the effective learning rate $\eta/b_{T}$ (c.f. Lemma \ref{lem: small_w_for_allbsq_sub}). This case shows that if $\alpha$ is relatively small that the second term is smaller than the first term, then we have the same rate as gradient descent. See Remark \ref{re:thm} for more discussion.

In the second case, the initial effective learning $\eta/b_0$ is greater than the threshold that guarantees the convergence of gradient descent.
Our algorithm will guarantee either of the followings happens after $T$ iterations.
(1) The loss is already small, so we can stop training.
This corresponds the first term $\frac{\left(\eta \|\vH^{\infty} \|\right)^2 -b_0^2 }{\alpha^2\sqrt{ n\varepsilon} } $.
(2) The loss is still large, which will make the effective stepsize $\eta/b_k$ decrease with a good rate. That is, if (2) keeps happening, the stepsize will decrease till $\eta/b_k \le 1/(C\norm{\mat{H}^\infty})$ and we are in the first case.
Note the first term is the same as the second term of the first case.
The third term $\left( \frac{\|\vH^\infty\|}{\lambda_0} \right)^{2}\log\left(\frac{1}{\epsilon}\right)$ is slightly worse than the rate in  the gradient descent. The reason is the loss may increase due to the large learning rate at the beginning. (c.f. Lemma \ref{lem:log_growth_sub}).   
%\simon{@Shirley,please add explaination where this term is from and refer a lemma. }  

To summarize, these two cases together show that our algorithm is robust to hyper-parameter choices. The bad choices of hyper-parameters will only hurt the constant in the convergence rate but the global polynomial time convergence is still guaranteed. 
%We  compare with gradient descent and give some concrete examples for our choice of  hyper-parameters below.

\begin{rem} \label{re:thm}
It is difficult to set the parameters with optimal values due to the fact that the maximum and minimum eigenvalues of the matrix $ \vH^{\infty}$  are computational costly and so generally unknown.  
According to Theorem~\ref{thm:main_gd}, since $n$ is an upper bound of $\norm{\mat{H}^\infty}$, one may  use  gradient descent by setting  $\eta = \Theta\left(\frac{1}{n}\right)$ and have the convergence rate of $T_1 = \widetilde{O}\left( \left(\frac{n}{\lambda_0} \right) \log \left( \frac{ 1}{ \varepsilon }\right) \right)$. 

%While for our algorithm if we set $b_0 /\eta = 2n$ and $ \alpha=\frac{\eta}{n}$, then we are most likely in the first case.  We can have the number of hidden layer $m$ same as gradient  descent and 
%\begin{align*}
%T_2 = O\left( \left(\frac{  n  }{  \lambda_0} +\frac{1}{n\lambda_0^2\sqrt{\delta}}\right) \log \left( \frac{ n}{ \varepsilon \delta }\right) \right),
%    \end{align*} 
%which almost  matches with $T_1$  of gradient descent. 
However, this choice of step size is not optimal when $ \|\vH^{\infty}\|$ is much smaller than  $n$. Using adaptive gradient algorithm with the small initialization on the effective learning rate  would results in better complexity. Indeed, for instance, let the target training error be $\varepsilon=\frac{1}{\sqrt{n}}$, the typical statistical target error and set $b_0 = \eta$, $ \alpha=\frac{1}{\sqrt{n}}$.  
Now in the scenario that $\|\vH^{\infty}\| = \Theta(1)$ and $ \frac{1}{\lambda_0}= \Theta\left(n^{9/8} \right)$, the convergence rate of our adaptive method is $T_2 = \widetilde{O}(n^{9/4})$ comparing to the convergence rate of gradient descent which is $T_1 = \widetilde{O}(n^{5/2})$.
\end{rem}

\subsection{Proof Sketch of Theorem~\ref{thm:main_adagradloss}}
\label{sec:adaloss_proof}
We prove by induction.
Our induction hypothesis is the following.
\begin{condition}
	\label{con:adagrad_sub}
	At the $k'$-th iteration,\footnote{ For the convenience of induction proof, we define 
		$ \norm{\vect{y}-\vect{u}(-1)}^2= \norm{\vect{y}-\vect{u}(0)}^2 /\left(1- \frac{\eta  \lambda_0 C_1}{b_{0}}\left(1-\frac{\eta C{\|\vH(0)\|}}{b_{0}} \right) \right). $
			} there exists a constant $C_1$ such that  \footnote{See Table \ref{table} in Appendeix \ref{sec:basic} for the expressions} 
	\begin{align}
	\norm{\vy-\vu(k')}^2 
	\leq  &\left(  1- \frac{\eta  \lambda_0C_1}{b_{k'}}\left(1-\frac{\eta C{\|\vH^{\infty}\|}}{b_{k'}} \right) \right)\norm{\vect{y}-\vect{u}(k'-1)}^2 \label{eq:decrease}.
	\end{align} 
\end{condition}
Recall the key Gram matrix $\mat H(k')$  at $k'$-th iteration 
\begin{align}
\mat{H}_{ij}(k’) = \frac{1}{m}\sum_{r=1}^{m}\vect{x}_i^\top \vect{x}_j \indict{\left\{\vect{w}_r(k')^\top \vect{x}_i \ge 0, \vect{w}_r(k')^\top \vect{x}_j \ge  0\right\}}. \label{eqn:Hk}  
 \end{align}
%The high level idea is the same as Theorem \ref{thm:main_gd} that Condition~\ref{con:adagrad_sub}  holds at $k+1$-th iteration  as long as $\{\vH(k')\}_{k'=0}^k$ are positive.  In this sense, our goal is to prove the positiveness of $\{\vH(k')\}_{k'=0}^k$ by induction.
We prove two cases   $b_0/\eta\geq C{\|\vH^{\infty}\|}$  and  $b_0/\eta\leq C{\|\vH^{\infty}\|}$ separately.
\paragraph{\textbf{Case (1)}: $b_0/\eta\geq C{\|\vH^{\infty}\|}$}  
The base case $k'=0$ holds by the definition.  Now suppose for $k'=0,\ldots,k$, Condition~\ref{con:adagrad_sub} holds and we want to show Condition~\ref{con:adagrad_sub} holds for $k'=k+1$.  
Because $b_0/\eta\geq C{\|\vH^{\infty}\|}$,  by Lemma \ref{lem: small_w_for_allbsq_sub} we have
\begin{align}
\|\vw_r(k)-\vw_r(0)\|
&\leq \frac{4 \sqrt{n} }{\sqrt{m} \lambda_0 C_1} 
 \norm{\vy-\vu(0)}.
 \label{eq:case1}
\end{align}
Next, plugging in $m =\Omega \left(\frac{n^6}{\lambda_0^4\delta^3} \right)$, we have $ \|\vw_r(k)-\vw_r(0)\|\leq \frac{c\lambda_0\delta }{n^2}$. Then by Lemma~\ref{lem:H0_close_Hinft} and\ref{lem:close_to_init_small_perturbation}, the matrix  $\vH(k)$ is positive such that the smallest eigenvalue of $\vH(k)$ is greater than $\frac{\lambda_0}{2}$. Consequently, we have Condition~\ref{con:adagrad_sub} holds for $k'=k+1$. 

Now we have proved the induction part.
Using Condition~\ref{con:adagrad_sub}, for any $T \in \mathbb{Z}^+$, we have 
% Now that the induction is proved, we iteratively substitute $ \norm{\vy-\vu(t)}^2, t = k-1, k-1, \cdots, 0$  in inequality \eqref{eq:decrease} and have
\begin{align*}
\norm{\vy-\vu(T)}^2 
   &\leq \Pi_{t=0}^{T-1}\left( 1-\frac{\eta \lambda_0C_1}{2 b_{t+1}}\right)\norm{\vy-\vu(0)}^2\\
  &\leq \exp\left( -T \frac{\eta \lambda_0C_1}{2b_{\infty}}  \right)\norm{\vy-\vu(0)}^2
\end{align*}
where $b_\infty = b_0 + \frac{4\alpha^2\sqrt{n}}{\eta \lambda_0 C_1}\norm{\vect{y}-\vect{u}(0)} = O(b_0 + \frac{\alpha^2n}{\eta^2 \lambda_0\sqrt{\delta}})$ (c.f. Lemma~\ref{lem: small_w_for_allbsq_sub}).
This implies the convergence rate of \textbf{Case (1)}.
%Plugging the upper bound $b_{\infty}$ of $b_k$ derived from Lemma \ref{lem: small_w_for_allbsq_sub}, we prove the theorem for $b_0/\eta\geq C{\|\vH^{\infty}\|}$.

\paragraph{\textbf{Case (2)}: $b_0/\eta \leq C{\|\vH^{\infty}\|}$}  
%Note if there exists some $k'$ such that $b_{k'}/\eta\geq 2 C{\|\vH^{\infty}\|}$ for some $k'$, Condition \ref{con:adagrad_sub} reduces to
%\begin{equation}
%\|\vy-\vu(k')\|^2 
%\leq  \left(  1- \frac{\eta  \lambda_0C_1}{2b_{k'}} \right)\norm{\vect{y}-\vect{u}(k'-1)}^2
%\label{eq:decrease}
%\end{equation}
%Suppose above inequality holds for all $k'$, we can prove there exits an upper bound for $b_k$ and so the convergence results follows. 
We define 
\[
\hat{T} = \arg\min_{k} \frac{b_k}{\eta} \ge C\norm{\mat{H}^\infty}.
\]
Note this represents the number of iterations to make \textbf{Case (2)} reduce to \textbf{Case (1)}.
We first give an upper bound $T_0$ of $\hat{T}$. If
\[
T_0 = \left\lceil \frac{ \left( \eta C{\|\vH^{\infty}\|} \right)^2- b_0^2 }{\alpha^2 \sqrt{ n\varepsilon }}  \right\rceil +1
\]
applying Lemma \ref{lem:increase_sub} 
with parameters $\gamma = \alpha^2 \sqrt{n} $,  $a_j = \|\vy-\vu(k)\|  $ and $L = \left( \eta C\|\vH^{\infty} \| \right)^2$
we have after $T_0$ step, 
\begin{align*}
\text{either } \quad \min_{k\in [{T}_0]} \|\vy-\vu(k)\|^2 \leq  \varepsilon, \quad  \text{or}\quad  \quad b_{{T}_0} \ge  \eta C\|\vH^{\infty} \| .
%\label{eq:twocond}
\end{align*}
If  $
\min_{k\in[T_0]}\norm{\vy-\vu(k)}^2\leq \varepsilon
$, we are done. 
%Therefore, $\widetilde{T}_0$ is an upper bound of $T_0$.
Note this bound $T_0$ incurs the first term of iteration complexity of the \textbf{Case (2)} in Theorem~\ref{thm:main_adagradloss}.
%In the rest of the proof we will assume $b_{T_0} \ge C\norm{\mat{H}^\infty}$.
%Thus, there is no concerns on whether  $b_{k}/\eta\geq  C{\|\vH^{\infty}\|}$ would be satisfied or not.

Similar to \textbf{Case (1)}, we use induction for the proof.
Again the base case $k'=0$ holds by the definition.  Now suppose for $k'=0,\ldots,k$, Condition~\ref{con:adagrad_sub} holds and we will show it also holds for $k'=k+1$. 
There are two scenarios.

For $k\leq T_0-1$, Lemma \ref{lem: small_w_for_allbsq1_sub} implies that $\|\vw_r(k)-\vw_r(0)\| $ is upper bounded.
% of  that updated by the bad learning rate $b_k/\eta\leq C{\|\vH^{\infty}\|}$ grows upto a finite number still controlled by $\frac{1}{\sqrt{m}}$. 
Now plugging in our choice on $m$ and using Lemma~\ref{lem:H0_close_Hinft} and~\ref{lem:close_to_init_small_perturbation}, we know $\lambda_{\min}\left(\vH(k)\right)\ge \lambda_0/2$ and $\norm{\mat{H}(k)} \le C\norm{\mat H^\infty}$.
These two bounds on $\mat H(k)$ imply Condition~\ref{con:adagrad_sub}.
% Thanks to large  $m$  such that  $\hat{R}\leq \frac{c\lambda_0\delta }{n^2}$, so as to have the positiveness of $\vH(k)$ like in Case (1).\simon{Put the exact lower bound here} Condition~\ref{con:adagrad_sub} holds for $k+1$, and then $k = T_0-1$ when $k= T_0-2$.

When  $k\geq T_0$, we have  contraction bound as in \textbf{Case (1)} and then same argument  follows but with the different initial values $\vW(T_0-1)$ and $\|\vy-\vu (T_0-1)\|$. 
We first analyze $\vW(T_0-1)$ and $\|\vy-\vu (T_0-1)\|$.
By Lemma~\ref{lem:log_growth_sub}, we know  $\|\vy-\vu (T_0-1)\|$  only increases an additive $ O\left(\left(  \eta C \| \vH^{\infty}\| \right)^{3/2}\right)$ factor from $\|\vy-\vu (0)\|$.
Furthermore, by Lemma~\ref{lem: small_w_for_allbsq1_sub}, we know for $r \in [m]$\begin{align*}
\norm{\vect{w}_r(T_0-1) - \vect{w}_r(0)} \le \frac{4\eta^2 C\norm{\mat{H}^\infty}}{\alpha^2\sqrt{m}}.
\end{align*}
Now we consider $k$-th iteration.
Applying Lemma~\ref{lem: small_w_for_allbsq_sub}, we have 
\begin{align*}
\|\vw_r(k)-\vw_r(0)\|
\leq &\|\vw_r(k)-\vw_r(T_0-1)\|+ \|\vw_r(T_0-1)-\vw_r(0)\|\\
\leq & \frac{4 \sqrt{n} }{\sqrt{m} \lambda_0 C_1} \left( 
\norm{\vy-\vu(T_0-1)} +\hat{R}
 \right)\\
 \leq  &{c\lambda_0\delta } /{n^2}
\end{align*}
where the last inequality we have used our choice of $m$.
Using Lemma~\ref{lem:H0_close_Hinft} and~\ref{lem:close_to_init_small_perturbation} again, we can show  $\lambda_{\min}\left(\vH(k)\right)\ge \lambda_0/2$ and $\norm{\mat{H}(k)} \le C\norm{\mat H^\infty}$.
These two bounds on $\mat H(k)$ imply Condition~\ref{con:adagrad_sub}.

Now we have proved the induction.
The last step is to use Condition~\ref{con:adagrad_sub} to prove the convergence rate.
Observe that for any $T \ge T_0$, we have 
\begin{align*}
& \norm{\vy-\vu(T)}^2 
\leq   \exp\left( - (T-T_0+1) \frac{\eta \lambda_0C_1}{2\bar{b}_\infty}  \right) \norm{\vy-\vu(T_0-1)}^2
\end{align*}
where we have used Lemma~\ref{lem: small_w_for_allbsq_sub} and Lemma~\ref{lem:log_growth_sub} to derive\begin{align*}
\bar{b}_\infty = \eta C\|\vH^{\infty}\|+\frac{4\alpha^2 \sqrt{n}}{\eta {\lambda_0}C_1}\left( \norm{\vy-\vu(0)} + \frac{ 2\eta^2 \sqrt{\lambda_0}   \left(C\|\vH^\infty\| \right)^{3/2}}{ \alpha^2 \sqrt{n}  } \right).
\end{align*}
With some algebra, one can show this bound corresponds to the second and the third term of iteration complexity of the \textbf{Case (2)} in Theorem~\ref{thm:main_adagradloss}.

%Plugging $
%T_0 = \lceil{ \frac{ \left( \eta C{\|\vH^{\infty}\|} \right)^2- b_0^2 }{\alpha^2 \sqrt{ n\varepsilon }}  \rceil}+1$,  the upper bound of $\norm{\vy-\vu(T_0-1)}^2$  from Lemma \ref{lem:log_growth_sub} and the upper bound $\bar{b}_{\infty}$ of $b_k$  derived from Lemma \ref{lem: small_w_for_allbsq_sub}  results in the statement for Case (2).
\subsubsection{Ingredients of Proof}
\label{subsec:ingredients}
As we have seen in the proof sketch. 
Lemma~\ref{lem: small_w_for_allbsq_sub} and Lemma~\ref{lem: small_w_for_allbsq1_sub} are most important lemmas in the proof of Theorem~\ref{thm:main_adagradloss}.
Here we state and prove these two lemmas.
 \begin{lem}
\label{lem: small_w_for_allbsq_sub}
Suppose Condition~\ref{con:adagrad_sub} holds for $k'=0,\ldots,k$ and   $ b_k$  is updated by Algorithm 1. Let $T_0\geq 1$ be the first index such that $b_{T_0} \geq  \eta C \|\vH^{\infty}\| $.  Then for every $r \in [m]$ and $k = 0,1,\cdots$,
\begin{align*}
 & b_{k}\leq b_{T_0-1}+\frac{4\alpha^2 \sqrt{n}}{\eta {\lambda_0}C_1}\norm{\vy-\vu(T_0-1)} ;\\
\|\vw_r(k+T_0)-&\vw_r(T_0-1)\| 
\leq \frac{4 \sqrt{n} }{\sqrt{m} \lambda_0 C_1} 
 \norm{\vy-\vu(T_0-1)}  \triangleq \widetilde{R} .
\end{align*}
\end{lem}

\textbf{Proof of Lemma \ref{lem: small_w_for_allbsq_sub}}
When $b_{T_0}/\eta \geq C \|\vH^{\infty}\|$ at some $T_0\geq 1$, thanks to the key fact that Condition~\ref{con:adagrad_sub} holds $k'=0,\ldots,k$, we have
\begin{align*}
\norm{\vy-\vu(k+T_0)} 
\leq & \sqrt{\left( 1-\frac{\eta \lambda_0 C_1}{2b_{k+T_0})} \right)}\norm{\vy-\vu(k+T_0-1)} \\
 \leq  &\left( 1-\frac{\eta \lambda_0 C_1}{4b_{k+T_0}} \right)\norm{\vy-\vu(k+T_0-1)} \\
 \leq & \norm{\vy-\vu(T_0-1)}-\sum_{t=0}^{k}\frac{\eta \lambda_0 C_1}{4b_{t+T_0}}\norm{\vy-\vu(t+T_0-1)} \\
 \Rightarrow  \quad \sum_{t=0}^{k} &\frac{\norm{\vy-\vu(t+T_0-1)}}{b_{t+T_0}} \leq  \frac{4 \norm{\vy-\vu(T_0-1)}   }{\eta \lambda_0 C_1}.
 \end{align*}
Thus, the upper bound for $b_k$,
\begin{align*}
b_{k+T_0} 
%=  & b_{k+T_0-1}  +\frac{\alpha^2\sqrt{m}}{b_{k+T_0}+b_{k+T_0-1}} \max_{r\in[m]}\norm{\frac{\partial L(\vW(k+T_0-1))}{\partial\vw_r}}_{2}\\
\leq& b_{k+T_0-1}  + \frac{\alpha^2\sqrt {n} }{b_{k+T_0}} \|\vy - \vu(k+T_0-1)\| \\
\leq&b_{T_0-1} +\sum_{t=0}^{k}\frac{\alpha^2\sqrt {n} }{b_{t+T_0}} \|\vy - \vu(t+T_0-1)\| \\
\leq &b_{T_0-1} +\frac{4\alpha^2\sqrt {n} }{\eta {\lambda_0}C_1}\norm{\vy-\vu(T_0-1)}.
\end{align*}
As for the  upper bound of $\|\vw_r(k+T_0)-\vw_r(T_0-1)\|,$
\begin{align*}
 \|\vw_r(k+T_0)-\vw_r(T_0-1)\|
\leq& \sum_{t=0}^{k}\frac{\eta }{b_{t+T_0}} \| \frac{\partial L(\vW{(t+T_0-1)})}{\partial \vw_r } \| \\
%\leq & \frac{\eta \sqrt{n}}{\sqrt{m}}\sum_{t=0}^{k}\frac{1}{b_{t+T_0}} \| \vy - \vu(t+T_0-1)\| \\
 \leq& \frac{4 \sqrt{n} }{\sqrt{m} \lambda_0 C_1} \norm{\vy-\vu(T_0-1)}.  
\end{align*}

\begin{lem}
\label{lem: small_w_for_allbsq1_sub}
Let  $T_0\geq 1$ be the first index such that $b_{T_0} \geq  \eta C \|\vH^{\infty}\| $. Then for every $r \in [m]$, we have for $k=0,1,\ldots,T_0-1$,
\begin{align*}
\|\vw_r(k)-\vw_r(0)\|
\leq 
 \frac{4\eta^2C\|\vH^{\infty}\|}{ \alpha^2\sqrt{m}} \triangleq \hat{R}.
\end{align*}
\end{lem}

\textbf{Proof of Lemma \ref{lem: small_w_for_allbsq1_sub}}
For the upper bound of $\|\vw_r(k+1)-\vw_r(0)\|$ when $b_{t}/\eta < C \|\vH^{\infty}\|$, $t = 0,1, \cdots, k$ and $k\leq T_0-2$, we first
observe that
\begin{align*}
 \sum_{t=0}^{k}\frac{\norm{\vy-\vu(t)}}{b_{t+1}}  
 &\leq  \frac{1}{ \alpha^2 \sqrt{ n} }\sum_{t=0}^{k}   \frac{ \alpha^2 \sqrt{ n} \norm{\vy-\vu(t)} }{  \sqrt{ \alpha^2\sqrt{ n}\sum_{\ell=0}^{t} \| \vy-\vu(\ell)\|+b_0^2}} \nonumber \\
&\leq  \frac{2}{ \alpha^2 \sqrt{ n} }\sqrt{  \alpha^2\sqrt{ n}\sum_{\ell=0}^{k}\| \vy-\vu(\ell)\|+b_0^2} \nonumber \\
&\leq  \frac{ 2b_{T_0-1} }{ \alpha^2 \sqrt{ n} }
\end{align*}
where the  second inequality use Lemma \ref{lem:sqrtsum} and the third inequality is due to the fact that $b_k \leq b_{T_0-1} \leq \eta C \|\vH^{\infty}\|$ for all $k\leq T_0-2$.  Thus,
\begin{align*}
\|\vw_r(k+1)-\vw_r(0)\|   \leq& \sum_{t=0}^{k}\frac{\eta }{b_{t+1}} \| \frac{\partial L(\vW{(t)})}{\partial \vw_r } \| 
\leq  \frac{\eta \sqrt{n}}{\sqrt{m}}\sum_{t=1}^{k}\frac{ \| \vy - \vu(t)\|}{b_{t+1}} 
\leq  \frac{2C \eta^2 \|\vH^{\infty}\|}{ \alpha^2\sqrt{ m}}.
\end{align*}

\section{Discussion on Variants of AdaGrad}   
\label{sec:diss}
In this section we compare our proposed algorithm AdaLoss with existing adaptive methods.
Algorithm  \ref{alg:adagradloss} can be viewed as a variant of the standard AdaGrad algorithm proposed by  \cite{duchi2011adaptive}, where the norm version of the update is
\begin{align*}
 b_{k+1}^2 = b_{k}^2 + \sqrt{m} \max_{r\in[m]} \bigg\| \frac{\partial L(\vW{(k)})}{\partial \vw_r } \bigg\|^2. 
\end{align*}
Our algorithm AdaLoss is similar to AdaGrad, but is distinctly different from AdaGrad: we update $b_{k+1}^2$ using the \emph{norm} of the \emph{loss} instead of the \emph{squared norm} of the \emph{gradient}.  We considered the AdaLoss update instead of AdaGrad because, in the setting considered here, the modifications allowed for dramatically better theoretical convergence rate. 

\paragraph{Why the Loss instead of the Gradient?}  Indeed, our update of $b^2_{k+1}$ is not too different from the following update rule using the gradient 
\begin{align}
b_{k+1}^2 = b_{k}^2 + \alpha^2\sqrt{m} \max_{r\in[m]} \bigg\| \frac{\partial L(\vW{(k)})}{\partial \vw_r } \bigg\|.
\label{eq:adagradss}
\end{align}
The AdaLoss update can be upper and lower  bounded   by $b_k^2$ and the norm of the gradient, i.e.,
\[   b_{k}^2 +  \alpha^2 \sqrt{m} \max_{r\in[m]}\bigg\| \frac{\partial L(\vW{(k)})}{\partial \vw_r } \bigg\|\leq b_{k+1}^2 \leq b_{k}^2 + \frac{ \alpha^2 \sqrt{m}}{\sqrt{\lambda_0}} \max_{r\in[m]}\bigg\| \frac{\partial L(\vW{(k)})}{\partial \vw_r } \bigg\| \]
where the first and second inequalities are respectively due to Proposition \ref{prop:a1} and  Proposition \ref{lem:grad_to_training_sub}. 
%\simon{Write a formal lemma and refers to it. Using \textbf{P1} looks weird to me}.
\begin{prop}
\label{lem:grad_to_training_sub}
If $\lambda_{min} (\vH)\geq \frac{\lambda_0}{2}$, then $\|\vy -\vu\| \leq \frac{ \sqrt{2m}}{\sqrt{\lambda_0}}\max_{r\in[m]}\| \frac{\partial L (\vW)}{\partial \vw_r}\|.$ \footnote{Proof  is given in Appendix \ref{sec:proof_prop} } 
\end{prop}
However, we use $\sqrt{n}\|\vy-\vu(k) \|$ instead of using the gradient to update $b_k$ because our convergence  analysis requires  lower and upper bounding the dynamics $b_1, \ldots, b_{k}$, in terms of $\|\vy-\vu(k) \|$. If $b_k$ were instead updated using  \eqref{eq:adagradss}, then 
%$b_{k+1}^2$ lower bounded by
$$b_{k+1}^2\geq b_{k}^2 + \frac{ \alpha^2}{\sqrt{2\lambda_0}} \max_{r\in[m]} \bigg\| \frac{\partial L(\vW{(k)})}{\partial \vw_r } \bigg\|.$$
The above lower bound of $b_k$ results in a larger $T$ in Case (2) by a factor of $\sqrt{n/\lambda_0}$.  Using the loss instead of the gradient to update $b_k$ is independently useful as reusing the already computed loss information for each iteration can save some computation cost and thus make the update more efficient.

\paragraph{Why the norm and not the squared-norm?}  For ease of comparison with Algorithm \ref{alg:adagradloss}, we switch from gradient information to loss and compare with two close variants: 
\begin{align}
b_{k+1}^2 = b_{k}^2 +  \alpha^2 \sqrt{n}\|\vy-\vu(k) \| ^2 \label{eq:adagradss1} \\
b_{k+1} = b_{k} + \alpha \sqrt{n}\|\vy-\vu(k) \| 
\label{eq:adagradss2}.
\end{align}
Equation \eqref{eq:adagradss1} using the ``square'' rule update  is the standarad AdaGrad proposed by  \cite{duchi2011adaptive} and has been widely recognized as important optimizer in deep learning -- especially for training sparse datasets. For our over-parameterized models, this update rule does give a better convergence result in Case 1 when $b_0/\eta \geq  C \|\vH^{\infty}\|$ \footnote{The convergence proof is straightforward and similar to the first case in Theorem \ref{thm:main_adagradloss}}. However,  when the initialization  $b_0/\eta\leq  C \|\vH^{\infty}\|$, we were only able to prove convergence in case the level of  over-parameterization (i.e., $m$) depends on the training error $1/\varepsilon$, the bottleneck resulting from the attempting to prove the analog of Lemma \ref{lem: small_w_for_allbsq1_sub} (see Proposition \ref{prop:adasq} below).

 \begin{prop}\label{prop:adasq}
Let  $T_0\geq 1$ be the first index such that $b_{T_0} \geq  \eta C \|\vH^{\infty}\| $. Consider the update of $b_{k}$ in \eqref{eq:adagradss1}.
Then for every $r \in [m]$, we have  for $k=0,1,\ldots,T_0-1$,
\begin{align*}
\|\vw_r(k+1)-\vw_r(0)\|_2
&\leq   \frac{\eta\sqrt{2(k+1)}}{\alpha^2 \sqrt{m}}\sqrt{  1+ 2\log \left(\frac{C \eta \|\vH^{\infty}\| }{b_0}\right)}.
\end{align*}
\end{prop}
On the other hand, the update rule in \eqref{eq:adagradss2} can resolve the problem because the growth of $b_k$ is larger than \eqref{eq:adagradss1} such that the upper bound of $\|\vw_r(k)-\vw_r(0)\|_2$ $k=0,1\ldots,T_0-1$, is better than that in Proposition \ref{prop:adasq} and even Lemma \ref{lem: small_w_for_allbsq1_sub} if $c<b_0<  \eta C \|\vH^{\infty}\|$ for some small $c$. However, the growth of $b_k$ remains too fast once the critical value of $\eta C \|\vH^{\infty}\|$ has been reached -- the upper bound $b_{\infty}$ we were able to show is exponential in $1/\lambda_0$ and also in the hyper-parameters $b_0$,$\eta$, $\alpha$ and $n$, resulting in an extremely large  $T$ compared to Case (2) in Thoeorem \ref{thm:main_adagradloss}.

%\input{numeric_adagrad.tex}

% Acknowledgments---Will not appear in anonymized version

\subsubsection*{Acknowledgments}
\label{sec:ack}
 This work was partially done while all 3 authors were with the Simons Institute for the Theory of Computing at UC Berkeley.  We thank the institute for the financial support and the organizers  of the program on ``Foundation of Data Science". We would also thank Facebook AI Research  for partial support of Rachel Ward's Research.

\bibliography{simonduref}
\bibliographystyle{plainnat}

\newpage
\appendix

\section{Experiments}
\label{sec:exp}
%\subsection{The Experiments}
We first plot the eigenvalues of the matrices $\{\mat{H} (k)\}_{k'=0}^{k}$  and then provide the details. 

\begin{figure}[ht!!!]
\centering
 \includegraphics[width=\linewidth]{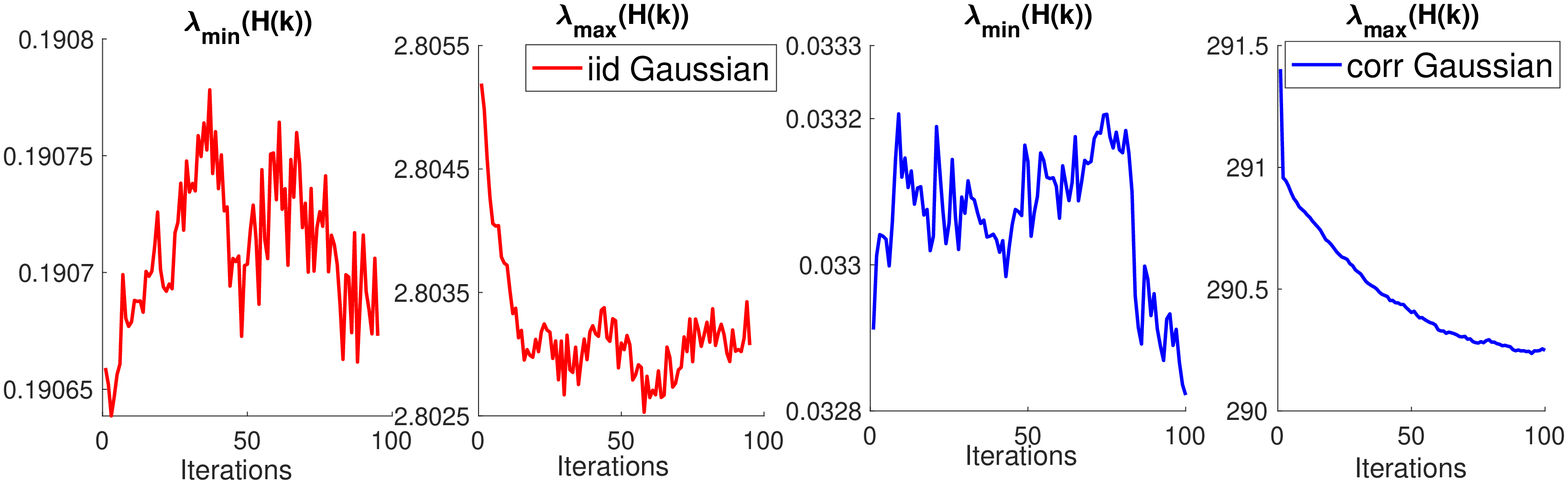}
 \includegraphics[width=\linewidth]{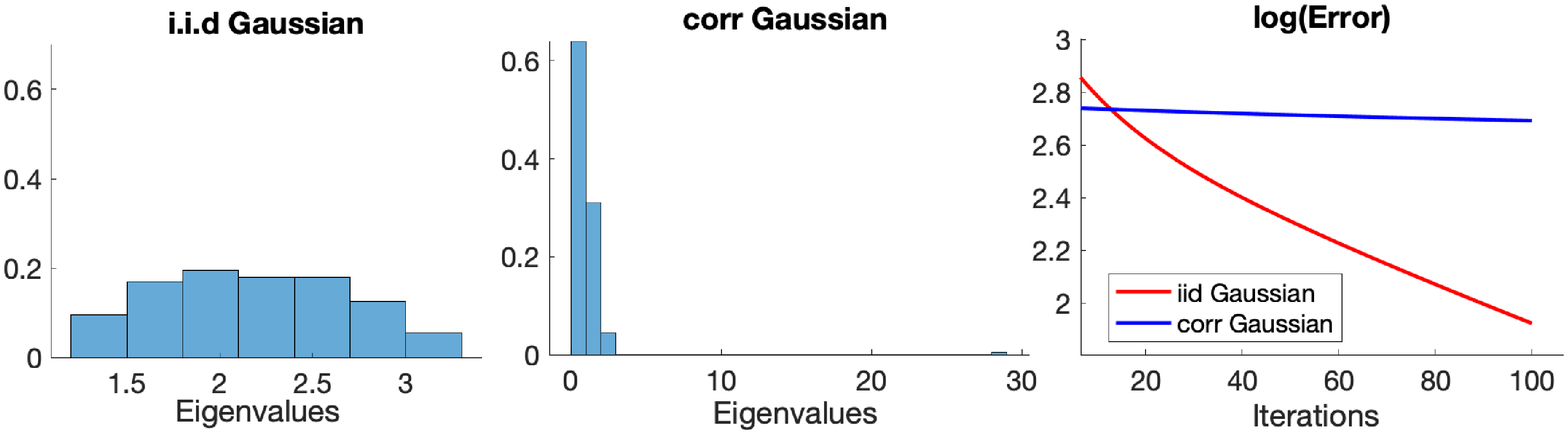}
\caption{ Top plots: y-axis is maximum or minimum eigenvalue of the matrix $\vH(k)$, x-axis is the  iteration. 
Bottom plots (left and middle): y-axis is the probability, x-axis is the  eigenvalue of co-variance matrix induced by Gaussian data.   Bottom plots (right): y-axis is the training error in logarithm scale, x-axis is the iteration. The distributions of eigenvalues for the co-variances matrix ($d \times d$ dimension) of the data  are  plotted on the left  for i.i.d. Gaussian and  in the middle  for  correlated Gaussian. The  bottom right plot  is the training error for the two-layer neural network $m=5000$ using the two Gaussian datasets.}\label{fig:data}
\end{figure}

We use  two simulated Gaussian data sets:  i.i.d. Gaussian (the red curves) and multivariate Gaussian (the blue curves).    Observe the red curves in Figure \ref{fig:data} that  the largest maximum eigenvalue is around $2.8$ and minimum eigenvalues is around $0.19$ within 100 iterations, while the  maximum  and minimum eigenvalues  for the blue curves  are  around $291$ and $0.033$ respectively.  To some extend,  i.i.d. Gaussian data  illustrates the case where the data points are  pairwise  uncorrelated  such that $ \norm{\mat{H}^{\infty}} =O\left(1\right)$, while correlated Gaussian data set implies the situation when the samples are highly correlated with each other $ \norm{\mat{H}^{\infty}} =O\left(n\right)$.

In the experiments, we simulate Gaussian data with training sample  $n=1000$  and  the dimension $d=200$. Figure \ref{fig:data} plots the histogram of  the eigenvalues of the co-variances for each dataset. Note that the eigenvalues are different from the eigenvalues in the top plots. We use  the two-layer neural networks $m=5000$. Although  $m$ here is far smaller than what Theorem \ref{thm:main_gd} requires, we found it sufficient for our purpose to just illustrate the maximum and minimum eigenvalues of  $\vH(k)$ for iteration $k = 0,1, \ldots, 100$. Set the learning rate  $\eta = 5\times 10^{-4}$ for i.i.d. Gaussian and  $\eta = 5\times 10^{-5}$ for correlated Gaussian. The training error is also  given in Figure \ref{fig:data}. 
%\subsection{Table}

\section{Proof for Theorem \ref{thm:main_gd}}
\label{sec:proof_gd}
We prove Theorem~\ref{thm:main_gd} by induction\footnote{Note that we use the same structure as in \cite{du2018gradient}. For the sake of completeness in the proof, we will use most of their lemmas, of which the proofs can be found  in technical section or otherwise in their paper .}.
Our induction hypothesis is the following convergence rate of empirical loss.
\begin{condition}\label{cond:linear_converge}
 At the $k$-th iteration,   we have  for $m =
\Omega\left(\frac{n^6}{\lambda_0^4\delta^3}\right)
$  such that with probability   $1-\delta$, $$\norm{\vect{y}-\vect{u}(k)}^2  \le (1-\frac{\eta \lambda_0}{2})^{k} \norm{\vect{y}-\vect{u}(0)}^2 .$$   
\end{condition}

\emph{Now we show Condition~\ref{cond:linear_converge} for every $k = 0,1,\ldots$.
For the base case $k=0$, by definition Condition~\ref{cond:linear_converge} holds.
Suppose for $k'=0,\ldots,k$, Condition~\ref{cond:linear_converge} holds and we want to show Condition~\ref{cond:linear_converge} holds for $k'=k+1$. We first prove the  order of $m$ and then the contraction of $\norm{\vect{u}(k+1)-\vect{y}}$.}
\subsection{ The order of $m$ at iteration $k+1$} \label{sec:order_m_gd}
Note that the contraction for $\norm{\vect{u}(k+1)-\vect{y}}$ is mainly controlled by the smallest eigenvalue of  the sequence of matrices $ \{\mat{H} (k')\}_{k'=0}^k$ . It requires that the minimum eigenvalues of matrix 
$\vH(k’), k'=0,1,\cdots, k$ are strictly positive, which is equivalent to ask that the update of $\vw_r(k')$ is not far away from initialization  $\vw_r(0)$ for $r\in[m]$. This requirement can be fulfilled by the large hidden nodes $m$. 

The first lemma (Lemma \ref{lem:H0_close_Hinft}) gives smallest $m$  in order to have $\lambda_{\min}(\mat{H}(0))>0$.
The next two lemmas concludes the order of $m$ so that  $\lambda_{\min}(\mat{H}(k'))>0$ for $k'=0,1,\cdots,k$. Specifically,  if $R' < R$, then the conditions in Lemma~\ref{lem:close_to_init_small_perturbation} hold for all $0\le k'\le k$.
We refer the proofs  of these lemmas to \cite{du2018gradient} 
\begin{lem}\label{lem:H0_close_Hinft}
If $m = \Omega\left(\frac{n^2}{\lambda_0^2} \log^2\left(\frac{n}{\delta}\right)\right)$, we have with probability at least $1-\delta$ that $\norm{\mat{H}(0)-\mat{H}^\infty} \le \frac{\lambda_0}{4}$.
\end{lem}

\begin{lem}\label{cor:dist_from_init}
If Condition~\ref{cond:linear_converge} holds for $k'=0,\ldots,k$, then we have for every $r \in [m]$
\begin{align}
\norm{\vect{w}_r(k'+1)-\vect{w}_r(0)} \le \frac{4\sqrt{n} \norm{\vect{y}-\vect{u}(0)}}{ \sqrt{m} \lambda_0} \triangleq R'.
\end{align}
\end{lem}
\begin{lem}\label{lem:close_to_init_small_perturbation}
Suppose for $r\in [m]$, $\norm{\vw_r - \vw_r(0)} \le \frac{c\lambda_0 \delta}{n^2 } \triangleq R$  for some small positive constant $c$.
Then we have with probability $1-\delta$ over initialization, $\norm{\mat{H} - \mat{H}(0)} \le \frac{\lambda_0}{4}$ where $\vH$ is defined in Definition \ref{def:H}.
\end{lem}
Thus it is sufficient to show $R' < R$. Since $\norm{\vect{y}-\vect{u}(0)}^2=O\left(\frac{n}{\delta} \right)$ derived from Proposition \ref{prop:a2},   $R' < R$ implies that \begin{align*}
m = & \Omega\left(\frac{n^5\norm{\vect{y}-\vect{u}(0)}^2 }{\lambda_0^4}\right)= \Omega\left(\frac{n^6 }{\lambda_0^4 \delta}\right).\\
\end{align*}
%By Markov's inequality, 
%$\norm{\vect{y}-\vect{u}(0)}^2  \leq O \left( \frac{n}{\delta} \right)$ with probability  at least $1-\delta$.

\subsection{The contraction  of  $\norm{\vect{u}(k+1)-\vect{y}}$}\label{sec:contraction_gd}
Define the event \[A_{ir} = \left\{\exists \vect{w}: \norm{\vect{w}-\vect{w}_r(0)} \le R, \indict\left\{\vect{x}_i^\top \vect{w}_r(0) \ge 0\right\} \neq \indict\left\{\vect{x}_i^\top \vect{w} \ge 0\right\} \right\} \quad \text{ with } 
R= \frac{c \lambda_0 \delta}{n^2}\]
for some small positive constant $c$.
%Recall in the proof of Theorem~ \ref{thm:main_gd}, we have $
%P(A_{ir}) = P_{z \sim N(0,1)}\left(\abs{z} < R\right) \le \frac{2R}{\sqrt{2\pi}}.
%$
%First, For every $i \in [n]$ and $r \in [m]$, we define the vent \[
%A_{ir} = \left\{\abs{\vect{w}_r(0)^\top \vect{x_i} }\ge R\right\}
%\]
%where we define \[
%R = \frac{c\lambda_0}{n^2}
%\] for some small positive constant $c$.
We let
$S_i = \left\{r \in [m]: \indict\{A_{ir}\}= 0\right\}$ and $
S_i^\perp = [m] \setminus S_i$.
 The following lemma bounds the sum of sizes of $S_i^\perp$.
\begin{lem}\label{lem:bounds_Si_perp}
With probability at least $1-\delta$ over initialization, we have $\sum_{i=1}^{n}\abs{S_i^\perp} \le  \frac{C_2mnR}{\delta}$  for some positive constant $C_2$.
\end{lem}

Next, we calculate the difference of predictions between two consecutive iterations,
\begin{align*}
[\vu(k+1)]_i - [\vu(k)]_i
= & \frac{1}{\sqrt{m}}\sum_{r=1}^{m} a_r \left(\relu{\left(
	\vect{w}_r(k) - \eta\frac{\partial L(\mat{W}(k))}{\partial \vect{w}_r(k)}
	\right)^\top \vect{x}_i} - \relu{\vect{w}_r(k)^\top \vect{x}_i}\right).
\end{align*}
Here we divide the right hand side into two parts.
$\vE_1^i$ accounts for terms that the pattern does not change and $\vE_2^i$ accounts for terms that pattern may change.

 Because $R' < R$, we know $\indict\left\{\vect{w}_r(k+1)^\top \vect{x}_i\ge 0\right\} \cap S_{i}= \indict\left\{\vect{w}_r(k)^\top \vect{x}_i \ge 0\right\}\cap S_{i}$.
\begin{align*}
[\vE_1]_i &\triangleq  \frac{1}{\sqrt{m}}\sum_{r \in S_i} a_r \left(\relu{\left(
	\vect{w}_r(k) - \eta\frac{\partial L(\mat{W}(k))}{\partial \vect{w}_r(k)}
	\right)^\top \vect{x}_i} - \relu{\vect{w}_r(k)^\top \vect{x}_i}\right) \\
	&= -\frac{1}{\sqrt{m}} \sum_{r \in S_i} a_r  \eta\langle{ \frac{\partial L(\mat{W}(k))}{\partial \vect{w}_r(k)}, \vx_i \rangle}\\
&=-\frac{\eta}{m}\sum_{j=1}^{n}\vect{x}_i^\top \vect{x}_j \left([\vu(k)]_j-\vy_j\right)\sum_{r \in S_i} \indict\left\{\vect{w}_r(k)^\top \vect{x}_i \ge 0, \vect{w}_r(k)^\top \vect{x}_j\ge 0 \right\}\\
&=  -\eta\sum_{j=1}^{n}([\vu(k)]_j-\vy_j)(\mat{H}_{ij}(k) -\mat{H}_{ij}^\perp(k))
\end{align*}
where $
\mat{H}_{ij}^\perp(k) = \frac{1}{m}\sum_{r \in S_i^\perp}\vect{x}_i^\top \vect{x}_j \indict\left\{\vect{w}_r(k)^\top \vect{x}_i \ge 0, \vect{w}_r(k)^\top \vect{x}_j \ge 0\right\}
$ is a perturbation matrix.
Let $\mat{H}^\perp(k)$ be the $n \times n$ matrix with $(i,j)$-th entry being $\vH_{ij}^\perp(k)$.
Using Lemma~\ref{lem:bounds_Si_perp}, we obtain  with  probability at least $1-\delta$, 
\begin{align} 
\norm{\mat{H}^\perp(k)} \le \sum_{(i,j)=(1,1)}^{(n,n)} \abs{\mat{H}_{ij}^\perp (k)}
\le \frac{n \sum_{i=1}^{n}\abs{S_i^\perp}}{m}
\le   \frac{n^2 mR}{\delta m} 
\le  \frac{n^2R}{\delta}.
\label{eq:norm_H_perp}
\end{align}
We view $[\vE_2]_i$ as a perturbation and bound its magnitude.
Because ReLU is a $1$-Lipschitz function and $\abs{a_r}  =1$, we have 
\begin{align}
[\vE_2]_i& \triangleq \frac{1}{\sqrt{m}}\sum_{r \in S_i^\perp} a_r \left(\relu{\left(
	\vect{w}_r(k) - \eta\frac{\partial L(\mat{W} (k))}{\partial \vect{w}_r(k)}
	\right)^\top \vect{x}_i} - \relu{\vect{w}_r(k)^\top \vect{x}_i}\right)\nonumber \\
&\le \frac{\eta }{\sqrt{m}} \sum_{r\in S_i^\perp} \norm{
	\frac{\partial L(\mat{W}(k))}{\partial \vect{w}_r(k)}
}  \nonumber\\
\Rightarrow   \quad  &\norm{\vE_2} \le  \frac{\eta \abs{S_i^\perp} \sqrt{n}\norm{\vect{u}(k)-\vect{y}}}{ m} 
%\nonumber \leq  \frac{\eta\sqrt{n}\norm{\vect{u}(k)-\vect{y}} }{m }\sqrt{\sum_{i=1}^n \abs{S_i^\perp}^2} 
 \leq \frac{\eta n^{3/2} R}{\delta} \norm{\vect{u}(k)-\vect{y}}\label{eq: I_2_upper}
\end{align}
Observe the maximum eigenvalue of matrix $\vH(k)$ upto iteration $k$ is bounded because
\begin{align}
\|\vH(k)-\vH(0)\| \leq \frac{4n^2R}{\sqrt{2\pi}\delta} \quad \text{with probability } 1-\delta \quad \Rightarrow \quad  \|\vH(k)\| \leq \|\vH(0)\| + \frac{4n^2R}{\sqrt{2\pi}\delta} \label{eq:upperH}
\end{align} 
Further, Lemma \ref{lem:H0_close_Hinft} \footnote{For more details, please see Lemma 3.1 in \cite{du2018gradient}} implies that 
\begin{align} 
\bigg|\|\vH^{\infty}\|-\|\vH(0)\|\bigg| = O\left( \frac{n^2\log(n/\delta)}{m} \right).
 \label{eq:upperH1}
\end{align}
That is, we could almost ignore the distance between $\|\vH^{\infty}\|$ and $\|\vH(0)\|$ for $m=\Omega\left(\frac{n^8}{\lambda_0^4\delta^3}\right)$.

With these estimates at hand, we are ready to prove the induction hypothesis.
\begin{align}
\norm{\vect{y}-\vect{u}(k+1)}^2 = 
&\norm{\vect{y}- \left(\vect{u}(k) + \vE_1 +\vE_2 \right)}^2 \nonumber\\
= &\norm{\vect{y}-\vect{u}(k)}^2 -  2\left(\vect{y}-\vect{u}(k)\right)^\top \left(\vE_1 +\vE_2\right) +\norm{\vE_1}^2 +\norm{\vE_2}^2+2\langle{\vE_1,\vE_2 \rangle} \nonumber\\
= &\norm{\vect{y}-\vect{u}(k)}^2 -   2\eta\left(\vect{y}-\vect{u}(k)\right)^\top \left(\mat{H}(k)-\mat{H}(k)^\perp\right) \left(\vect{y}-\vect{u}(k)\right) - 2 \left(\vect{y}-\vect{u}(k)\right)^\top\vect{E}_2 \nonumber \\
&+\eta^2\left(\vect{y}-\vect{u}(k)\right)^\top \left(\mat{H}(k)-\mat{H}(k)^\perp\right) ^2 \left(\vect{y}-\vect{u}(k)\right)  + \|\vect{E}_2\|^2 +2\langle{\vE_1,\vE_2 \rangle}\nonumber\\
= &\underbrace{\norm{\vect{y}-\vect{u}(k)}^2  -   2\eta\left(\vect{y}-\vect{u}(k)\right)^\top\left(\mathbf{I}- \frac{\eta}{2}\mat{H}(k)\right) \mat{H}(k) \left(\vect{y}-\vect{u}(k)\right)}_{{Term 1}} \nonumber\\
&+  2\eta\underbrace{\left(\vect{y}-\vect{u}(k)\right)^\top\left(  \frac{\eta}{2}\mat{H}^\perp(k) - \mathbf{I} \right) \mat{H}^\perp(k) \left(\vect{y}-\vect{u}(k)\right)}_{{Term 2}} \nonumber\\
&+  \underbrace{ \langle{2\vect{E_1} +\vect{E_2}-2 \left(\vect{y}-\vect{u}(k)\right) , \vect{E}_2 \rangle} }_{{Term 3}}+ 2\eta^2 \underbrace{\left(\vect{y}-\vect{u}(k)\right)^\top\mat{H}(k)\mat{H}^\perp(k) \left(\vect{y}-\vect{u}(k)\right)}_{{Term 4}}  
\label{eq:3terms}
\end{align}
Note that  $Term1$ dominates the increase or decrease of the term $\norm{\vect{y}-\vect{u}(k+1)}^2$ and other terms are very small for significant large $m$. 

First, given the strictly positiveness of matrix $\vH(k)$ and  the range of stepsize such that $\eta \leq \frac{1}{\| \mat{H}(k) \| }$,  we have
\begin{align*}
Term 1& \leq \left( 1-2\eta\left(1-\frac{ \eta}{2} ||\mat{H}(k)\| \right) \lambda_{min}(\mat{H}(k)) \right)\norm{\vect{y}-\vect{u}(k)}^2 
%& \leq \left( 1- \norm{\vect{y}-\vect{u}(k)}^2 \geq \frac{\lambda_0}{2} \left(1-\frac{ \eta}{2}||\mat{H}(k)\| \right) \right
\end{align*}
Due to \eqref{eq:norm_H_perp}, we could bound  $Term2$ 
\begin{align*}
Term 2 \leq \left(\frac{ \eta}{2} \|\mat{H}^\perp(k)\| ^2+\|\mat{H}^\perp(k)\| \right) \norm{\vect{y}-\vect{u}(k)}^2 \leq  \left( \frac{\eta n^2R}{2\delta}  + 1  \right) \frac{n^2R }{\delta}  \norm{\vect{y}-\vect{u}(k)}^2
\end{align*}
Due to \eqref{eq: I_2_upper}, we could bound  $Term 3$
\begin{align*}
Term 3& \leq  \left(2\|\vect{E_1}\|+ \|\vect{E_2}\|  +2 \norm{\vect{y}-\vect{u}(k)} \right) \|\vect{E_2}\|\\
& \leq \left( 2\eta \left( \|\mat{H}(k) \| +\| \mat{H}^\perp(k)\| \right)  \|\vy-\vu(k)\|+ \|\vect{E_2}\| + 2 \|\vy-\vu(k)\| \right) \|\vect{E_2}\|\\
& \leq \left( 2 \eta \| \mat{H}(k)\|+\frac{2\eta n^2R }{\delta}+1+ \frac{2\eta  n^{3/2} R}{\delta}  \right) \frac{\eta n^{3/2} R}{\delta} \norm{\vect{u}(k)-\vect{y}}^2 
\end{align*}
Finally, for $Term 4$
\begin{align*}
Term 4 \leq \|\mat{H}(k) \| \| \mat{H}^\perp(k)\|\norm{\vect{u}(k)-\vect{y}}^2 \leq \frac{n^2R}{\delta}  \|\mat{H}(k) \| \norm{\vect{u}(k)-\vect{y}}^2
\end{align*}

Putting $Term1$,  $Term2 $,  $Term3$  and  $Term4$ back to inequality \eqref{eq:3terms}, we have with probability $1-\delta$
\begin{align}
&\norm{\vect{y}-\vect{u}(k+1)}^2 -\norm{\vect{y}-\vect{u}(k)}^2 \nonumber\\
\leq & - 2\lambda_0 \eta \left( 1-   \frac{n^{3/2}\left( \sqrt{n}+2 \right)R }{2\lambda_0 \delta}- \frac{\eta}{2} \left( \frac{ \left(\lambda_0  \delta +2 n^{3/2}R+ n^2R\right)\|\vH(k)\|}{ \lambda_0  \delta} +\delta_1\right)\right)\norm{\vect{y}-\vect{u}(k)}^2
% \\
%\leq &- 2\eta \lambda_0C_1 \left(1-\frac{\eta}{2} C{\|\vH^{\infty}\|}\right) \norm{\vect{y}-\vect{u}(k)}^2
\label{eq:keyforadagrad1}
\end{align}
where $\delta_1= \left(\frac{n^{5/2}R }{2\delta}+\frac{2 n^2R }{\delta}+ \frac{2 n^{3/2} R}{\delta}  \right)\frac{n^{3/2} R}{\delta} $. Recall that $R= \frac{c\lambda_0 \delta}{n^3}$ for very small constant $c$; 
let  $C_1 = 1-   \frac{n^{3/2}\left( \sqrt{n}+2 \right)R }{2\lambda_0 \delta} = $ and 
 $C  = \frac{ { \left(\lambda_0  \delta +2 n^{3/2}R+ n^2R\right)\left( \|\vH^{\infty}\| + \frac{4n^2R}{\sqrt{2\pi}\delta} \right)}/{ \lambda_0  \delta} +\delta_1 }{ C_1 \| \vH^{\infty}\|} >0$. Since the upper bound of $\vH(k)$ is $  \|\vH^{\infty}\| + \frac{4n^2R}{\sqrt{2\pi}\delta}$ due to  \eqref{eq:upperH} and \eqref{eq:upperH1}, then we can re-write  \eqref{eq:keyforadagrad1} as follows
  \begin{align}
&\norm{\vect{y}-\vect{u}(k+1)}^2 -\norm{\vect{y}-\vect{u}(k)}^2 \nonumber\\
\leq &- 2\eta \lambda_0C_1 \left(1-\frac{\eta}{2} C{\|\vH^{\infty}\|}\right) \norm{\vect{y}-\vect{u}(k)}^2
\label{eq:keyforadagrad}
\end{align}
 
 We  have contractions for $\norm{\vect{y}-\vect{u}(k+1)}^2 , $  if the stepsize satisfy 
\begin{align*}
\eta \leq  \frac{2}{C{\|\vH^\infty\|}}  \quad \Rightarrow  \quad \eta   \leq \frac{1}{\|\vH(k)\|} . 
%\label{eq:stepsize}
\end{align*}
 We could pick  $\eta = \frac{1}{C{\|\vH^\infty\|}} =\Theta\left( \frac{1}{\|\vH^{\infty}\|}\right) $ for  large $m$ such that
\begin{align}
\norm{\vect{y}-\vect{u}(k+1)}^2 
& \leq  \left(1-\frac{\lambda_0C_1}{C{\|\vH^\infty\|}}\right)\norm{\vect{y}-\vect{u}(k)}^2 .  \nonumber
\end{align}
Therefore Condition~\ref{cond:linear_converge} holds for $k'=k+1$. Now by induction, we prove Theorem~\ref{thm:main_gd} .

\section{Proof for Theorem \ref{thm:main_adagradloss}}
\label{sec:proof_adaloss}
%\simon{Suggestion on proof structure:
%\begin{itemize}
%	\item A subsection just proving $\frac{b_0}{\eta} \ge C \norm{\mat{H}^\infty}$. Use induction as the current form and bound on $b_\infty$.
%	\item A subsection for $\frac{b_0}{\eta} \ge C \norm{\mat{H}^\infty}$. Use the same induction, add bounds on $T_0$ and $b_\infty$.
%\end{itemize}	
%	}
%
%

 In this section, we give the detailed proof for Theorem \ref{thm:main_adagradloss}.
The proof are organized into two parts. \textbf{Part I} in Subsection \ref{part1} is to prove the convergence for the initialization  $b_0/\eta \geq C \|\vH^{\infty}\|$. \textbf{Part II} in Subsection \ref{part2}  is to prove the convergence for the initialization  $b_0/\eta< C \|\vH^{\infty}\|$.   Several key lemmas will be stated and used during the proof,  and the proof of these lemmas will be deferred to subsection \ref{part3}. 

\subsection{Part I}
\label{part1}
In this part, we  prove the following condition by induction: \emph{show Condition~\ref{con:adagrad_sub} for every $k =0,1,2\ldots$. Based on this condition, we then obtain the upper bound of $b_k$ and so the convergence result.}\\
\textbf{Condition \ref{con:adagrad_sub}}
At the $k$-th iteration,  
\begin{align}
 \norm{\vy-\vu(k)}^2 
   &\leq \left(  1- \frac{\eta  \lambda_0 C_1}{b_{k}}\left(1-\frac{\eta C{\|\vH^\infty\|}}{b_{k}} \right) \right)\norm{\vect{y}-\vect{u}(k-1)}^2 
\label{eq:main_adagrad1}
\end{align} 
where we define 
$\norm{\vect{y}-\vect{u}(-1)}^2 =\norm{\vect{y}-\vect{u}(0)}^2  /\left(1- \frac{\eta  \lambda_0 C_1}{b_{0}}\left(1-\frac{\eta C{\|\vH^\infty\|}}{b_{0}} \right) \right) $,   $C_1$ and $C$ are some constant  of order $1$ (see Table \ref{table}  in Appendix \ref{sec:basic} for details).

For the base case $k'=0$, by definition Condition~\ref{con:adagrad_sub} holds. Suppose for $k'=0,\ldots,k$, Condition~\ref{con:adagrad_sub} holds and we want to show it still holds for $k'=k+1$. We first prove the  order of $m$ and then the contraction of $\norm{\vect{u}(k+1)-\vect{y}}$.

For the order of $m$ that controls the strict positiveness of $\{\vH(k')\}_{k'=0}^k$ and so the contraction of $\norm{\vect{u}(k+1)-\vect{y}}$, \footnote{For more descriptions of the relationship between $m$ and $\norm{\vect{u}(k+T_0)-\vect{y}}$, we refer to Subection \ref{sec:order_m_gd}} we  first have that $m = \Omega\left(\frac{n^2}{\lambda_0^2} \log^2\left(\frac{n}{\delta}\right)\right)$  from Lemma \ref{lem:H0_close_Hinft}. 
Further, by Lemma \ref{lem: small_w_for_allbsq_sub} with $T_0=1$ for the adaptive stepsize $b_{k'}\geq \eta C\|\vH^{\infty}\|$ , we can upper bound $\|\vw_r(k+1)-\vw_r(0)\|$ by  $ \widetilde{R}$. By Lemma \ref{lem:close_to_init_small_perturbation}, we ask for $\widetilde{R}\leq R$ for $k'=0,1,\cdots, k+1$, which results in  $m = \Omega(\frac{n^6}{\lambda_0^4\delta^3}). $ \\ 
\textbf{Lemma \ref{lem: small_w_for_allbsq_sub}}
Suppose Condition~\ref{con:adagrad_sub} holds for $k'=0,\ldots,k$ and  $ b_k$ is updated by Algorithm 1. Let  $T_0\geq 1$ be the first index such that $b_{T_0} \geq  \eta C \|\vH^{\infty}\|$. Then for every $r \in [m]$, we have for $k = 0,1,\cdots,$
\begin{align*}
 & b_{k}\leq b_{T_0}+\frac{4\alpha^2 \sqrt{n}}{\eta {\lambda_0}C_1}\norm{\vy-\vu(T_0-1)};\\
\|\vw_r(k+T_0)-&\vw_r(T_0-1)\| 
\leq \frac{4 \sqrt{n} }{\sqrt{m} \lambda_0 C_1} 
 \norm{\vy-\vu(T_0-1)} \triangleq \widetilde{R}.
\end{align*}

Now given the strictly positiveness of $\{\vH(k')\}_{k'=0}^k$ such that $\lambda_{min}\left( \vH(k')\right)>\frac{\lambda_0}{2}$,, we will prove \eqref{eq:main_adagrad1} at iteration $k+1$. We follow the same argument as Subsection  \ref{sec:contraction_gd} and straightforwardly modify the constant learning rate $\eta$ for the adaptive learning rate $\eta/b_{k+1}$. Observe the key inequality \eqref{eq:keyforadagrad} in Subsection  \ref{sec:contraction_gd} that  expresses the gradient descent with constant learning rate $\eta$ in Theorem \ref{thm:main_gd} and we have
\begin{align}
 \norm{\vy-\vu(k+1)}^2 
   &\leq \left(  1- \frac{\eta  \lambda_0 C_1}{b_{k+1}}\left(1-\frac{\eta}{b_{k+1}} C_{\|\vH(0)\|}\right) \right)\norm{\vect{y}-\vect{u}(k)}^2 
  \label{eq:LipchitzFuc}
\end{align} 
which is Condition \ref{con:adagrad_sub} for $k'=k+1$.

Now that we have proved Condition \ref{con:adagrad_sub} for all $k=0,1,2,\cdots,$ when $b_0/\eta \geq C\|\vH^{\infty} \|$. We use  Lemma \ref{lem: small_w_for_allbsq_sub} again to bound $b_k$ denoted by $b_{\infty}$:
$$b_{\infty} \leq  b_{0}+\frac{4\alpha^2 \sqrt{n}}{\eta {\lambda_0}C_1}\norm{\vy-\vu(0)}=O\left(b_{0}+\frac{\alpha^2 n}{\eta \lambda_0 \delta}  \right) $$
where the equality is from Proposition \ref{prop:a2}. 
%\begin{lem}
%\label{lem:bupper}
%Suppose Condition~\ref{con:adagrad} holds for $k'=0,\ldots,k$ and  $ b_k$  updated by Algorithm 1 . Let  $T_0\geq 1$ be the first index such that $b_{T_0} /\eta \geq  C \|\vH^{\infty}\|$. Then $b_k$ is bounded by 
%\begin{align*}
%b_{k}  \leq b_{T_0}+\frac{2\alpha^2 \sqrt{n}}{\eta {\lambda_0}C_1}\norm{\vy-\vu(T_0-1)},\quad  k = 0,1,\cdots.
%\end{align*}
%\end{lem} 
Thus, iteratively substituting  $ \norm{\vy-\vu(t)}^2, t = k-1, k-1, \cdots, 0$  in inequality \eqref{eq:LipchitzFuc}, we have
\begin{align}
 \norm{\vy-\vu(T)}^2 
  &\leq \Pi_{t=0}^{T-1}\left( 1-\frac{\eta \lambda_0C_1}{2 b_{t}}\right)\norm{\vy-\vu(0)}^2
%  &\leq \exp\left( - \sum_{t=0}^{T-1}\frac{\eta \lambda_0C_1}{2b_{t}} \right)\norm{\vy-\vu(0)}^2 \nonumber\\
  \leq \exp\left( -T \frac{\eta \lambda_0C_1}{2b_{\infty}}  \right)\norm{\vy-\vu(0)}^2 \nonumber.
\end{align}
For tolerance error $\varepsilon$ such that $ \norm{\vy-\vu(T)}^2 \leq \varepsilon$, we get the maximum step by plugging the upper bound $b_{\infty}$ into above inequality.

%Since $\{b_{t}\}_{t=0}^{k+1}$ is a monotone sequence, we have $b_{k+T_0}\geq b_{k+T_0-1}\geq b_{T_0} \geq L = \max\{1, 2\eta C_{\| \mat{H}(0)\|} \}$, thus  inequality \eqref{eq:LipchitzFuc} can be further reduced to
%\begin{align*}
% \norm{\vy-\vu(k+T_0)}^2 
%   &\leq \left(  1- \frac{\eta \lambda_0 C_1}{2b_{k+T_0}} \right)\norm{\vect{y}-\vect{u}(k+T_0-1)}^2  \\
%      &\leq \left(  1- \frac{\eta \lambda_0 C_1}{2b_{k+T_0}} \right) \Pi_{t=0}^{k-1}(1-\frac{\eta \lambda_0}{2b_{t+T_0}})\norm{\vy-\vu(T_0-1)}^2 \\
%           &\leq  \Pi_{t=0}^{k}(1-\frac{\eta \lambda_0}{2b_{t+T_0}})\norm{\vy-\vu(T_0-1)}^2 
%\end{align*} 

\subsection{Part II}
\label{part2}

Starting with $ b_0/\eta < C\|\vH^{\infty} \|$, we  use Lemma \ref{lem:increase_sub} 
with $\gamma = \alpha^2 \sqrt{n} $,  $a_j = \|\vy-\vu(k)\|  $ and $L = \left( \eta C\|\vH^{\infty} \| \right)^2$
 to prove that eventually after step 
\begin{align}
T_0=\bigg \lceil{ \frac{\left(\eta C\|\vH^{\infty} \|\right)^2 -b_0^2 }{\alpha^2 \sqrt{ n \varepsilon} }  \bigg \rceil} +1,
\label{eq:T0}
\end{align}
we have
\begin{align}
\text{either } \quad \min_{t\in [T_0]} \|\vy-\vu(t)\|^2 \leq  \varepsilon, \quad  \text{or}\quad  \quad b_{T_0} \ge  \eta C\|\vH^{\infty} \| .
\label{eq:twocond}
\end{align}
\textbf{Lemma \ref{lem:increase_sub}}
Fix $\varepsilon \in (0,1]$, $L > 0$, $\gamma>0$.   For any non-negative $a_0, a_1, \dots, $ the dynamical system
$$
b_0 > 0; \quad  \quad b_{j+1}^2 = b_j^2 +  \gamma a_j
$$
has the property that after 
{$N = \lceil{ \frac{ L^2-b_0^2}{ \gamma\sqrt{\varepsilon}} \rceil}+1$}
iterations, either $\min_{k=0:N-1}  a_k  \leq  \sqrt{\varepsilon}$, or $b_{N} \geq  L$. 

\emph{Now similar to \textbf{ Part I}, we  first use induction to prove Condition \ref{con:adagrad_sub} with $m$ satisfying 
\begin{align}
m&=\Omega \left ( \frac{4 n^{5} }{\lambda_0^4\delta^2 }  
 \left(  2\norm{\vy-\vu(0)}^2 + \frac{ \eta^4C^2 \|\vH^\infty\|^2}{ 2\alpha^4n } \right)\right) . 
\label{eq:m_large}
\end{align}
Note that since $T_0> 1$,  we will first prove the induction before $k\leq T_0-1$ and then  $k\geq T_0$. Based on this condition, we then obtain the upper bound of $b_k$ and so the convergence result.}

 For  $k'=0$, by definition Condition~\ref{con:adagrad_sub} holds. Suppose for $k'= 0, 1,\ldots, k \leq T_0-2$ , Condition~\ref{con:adagrad_sub} holds and we want to show it  holds for $k'=k+1 \leq T_0-1$. 
 Similar to \textbf{Part I}, we use Lemma \ref{lem:close_to_init_small_perturbation} in order to maintain the strict positiveness of $\vH(k
 )$  for $k'=k+1 \leq T_0-1$. That is,   we ask for $k'=0,1,\cdots, k+1$ such that  $ \|\vw_r(k')-\vw_r(0)\|\leq R= \frac{c\lambda_0 \delta}{n^2}$.  From Lemma~\ref{lem: small_w_for_allbsq1_sub}, we know that the upper bound of the distance $\|\vw_r(t)-\vw_r(0)\|$ for $t\leq T_0-1$ grows  only upto  a finite number  proportional to $ \eta C \| \vH^{\infty}\|$.
\\
\textbf{Lemma \ref{lem: small_w_for_allbsq1_sub}}
Let $L = \eta C \|\vH^{\infty}\|$  and $T_0\geq 1$ be the first index such that $b_{T_0} \geq  L $. Then for every $r \in [m]$, we have $k\leq T_0-2$
\begin{align*}
\|\vw_r(k+1)-\vw_r(0)\|
&\leq 
 \frac{2\eta^2 C\|\vH^{\infty}\|}{ \alpha^2\sqrt{ m}} \triangleq \hat{R}.
\end{align*}

Thus, we have  the strict positiveness of $\{\vH(k')\}_{k'=0}^k,, k \leq T_0-1$ such that $\lambda_{min}\left( \vH(k')\right)>\frac{\lambda_0}{2}$, as long as $ \hat{R}\leq \frac{c\lambda_0 \delta}{n^2}$ holds. But that is guaranteed for large  $m$ satisfying \eqref{eq:m_large}.
Again, use the same argument as the derivation of inequality \eqref{eq:LipchitzFuc}, we have Condition \ref{con:adagrad_sub} holds for $k'=k+1$.
Thus we prove Condition \ref{con:adagrad_sub} for $k'=0,1,\ldots, T_0-1$.

Now we are at $k' = T_0$. We have no clue at iteration $k'=T_0$ since there is no contraction bound  for  $\norm{\vy-\vu(k')}^2$,  $k'=0,1, k \leq T_0-1$. However, we have \eqref{eq:twocond} at $k'=T_0$. 

If we are lucky to have $$ \min_{t\in [T_0]} \|\vy-\vu(t)\|^2 \leq  \varepsilon,$$ then we are done. Otherwise, we have $b_{T_0}\geq \eta C\|\vH^\infty\|$. We will continue to prove the Condition \ref{con:adagrad_sub} for $k'=T_0, T_0+1,\ldots$. 

For the case $k'=T_0$, we need to prove the  strictly positive $\vH(T_0)$ given  Condition \ref{con:adagrad_sub} holds for $k'= 0, 1, \ldots, T_0-1$. At this time,  the upper bound of $ \norm{\vy-\vu(T_0-1)}  $ only grows up to a  factor  of $\eta C\|\vH^{\infty}\|$ as stated in following lemma,
\begin{lem}
\label{lem:log_growth_sub}
Let $T_0\geq 1$ be the first index such that $b_{T_0} \geq  \eta C \|\vH^{\infty}\| $. Suppose Condition \ref{con:adagrad_sub} holds for $k=0,1,\ldots,k$. Then
\begin{align*}
\norm{\vy-\vu(T_0-1)}
 \leq & \norm{\vy-\vu(0)} +  \frac{ 2\eta^2  \left(C\|\vH^\infty\|\right)^{2}}{   \alpha^2\sqrt{n}}.
\end{align*}
\end{lem}
Then the distance between $\vw_r(T_0)$ and $\vw_r(0)$  is
\begin{align*}
\|\vw_r(T_0)-\vw_r(0)\|
&\leq \|\vw_r(T_0)-\vw_r(T_0-1)\|+ \|\vw_r(T_0-1)-\vw_r(0)\| \nonumber\\
&\leq  \frac{\eta}{b_{T_0}} \norm{\frac{\partial L(\vW{(T_0-1)})}{\partial \vw_r}} + \frac{2\eta^2 C\|\vH^{\infty}\|}{ \alpha^2\sqrt{ m}} \nonumber\\
&\leq  \frac{ \sqrt{n} \|\vy-\vu(T_0-1)\| }{C\|\vH^{\infty} \| \sqrt{m}}  + \frac{2\eta^2 C\|\vH^{\infty}\|}{ \alpha^2\sqrt{ m}} \\
&\leq  \frac{1}{\sqrt{m}} \left(  \frac{ \sqrt{n} \|\vy-\vu(0)\| }{C\|\vH^{\infty} \| }  + \frac{4\eta^2}{ \alpha^2} C\|\vH^{\infty}\| \right)\\
&\leq   \frac{c\lambda_0\delta}{n^2}
\end{align*}
where the last inequality is due to large $m$ satisfying equation \eqref{eq:m_large}. Thus Lemma \ref{lem:close_to_init_small_perturbation}  implies that $\vH(T_0)$ is strict positive such that $\lambda_{min}(\vH(T_0))\geq \frac{\lambda_0}{2}>0$ and so Condition \ref{con:adagrad_sub} holds for $k=T_0$.
%since the lower bound of $b_{T_0-1}$ is 
%\begin{align*}
%b_{T_0-1} &= b_{T_0-2} +\frac{ \alpha^2 \sqrt{n}\norm{\vy-\vu(T_0-2)} }{b_{T_0-1}+ b_{T_0-2}}\\
%&\geq  b_{T_0-2} +\frac{ \alpha^2 \sqrt{n}\norm{\vy-\vu(T_0-1)} }{2b_{T_0-1}} \\
%&\geq  b_{0} +\sum_{t=0}^{T_0-1} \frac{ \alpha^2 \sqrt{n}\norm{\vy-\vu(T_0-1)} }{2b_{T_0-1}}
%\end{align*}

Now suppose for $k'= 0,\ldots, T_0-1, T_0,\ldots,k+T_0-1,$  Condition~\ref{con:adagrad_sub} holds and we want to show it holds for $k'=k+T_0$. The bound $\|\vw_r(k+T_0)-\vw_r(0)\|$  can be obtained by Lemma \ref{lem: small_w_for_allbsq_sub} and Lemma \ref{lem: small_w_for_allbsq1_sub}
\begin{align}
\|\vw_r(k+T_0)-\vw_r(0)\|
&\leq \|\vw_r(k+T_0)-\vw_r(T_0-1)\|+ \|\vw_r(T_0-1)-\vw_r(0)\| \nonumber\\
&\leq  \frac{4 \sqrt{n} }{\sqrt{m} \lambda_0 C_1} \left( 
\norm{\vy-\vu(T_0-1)} + \frac{ \eta^2\lambda_0 C\|\vH^{\infty}\| }{ 2\alpha^2 \sqrt{n}} \right)
 .\label{eq:mLa}
\end{align}
Putting back the upper bound of $\norm{\vy-\vu(T_0-1)}$ given in Lemma \ref{lem:log_growth_sub} and using Lemma \ref{lem:close_to_init_small_perturbation} that ask for $ \|\vw_r(k+T_0)-\vw_r(0)\|\leq R= \frac{c\lambda_0 \delta}{n^2}$, we require following
\begin{align*}
 \frac{c\lambda_0 \delta}{n^2} &\geq   \frac{4 \sqrt{n} }{\sqrt{m} \lambda_0 C_1} \left(  \norm{\vy-\vu(0)} + \frac{ \eta^2 C\|\vH^\infty\|(C\|\vH^\infty\|+\lambda_0)}{ 2\alpha^2 \sqrt{n}} \right).
\end{align*}
Rearranging  $m$ to one side, we get \eqref{eq:m_large}.
Given the strictly positive $\{\vH(k')\}_{k'=0}^{k+T_0}$,  we have Condition \ref{con:adagrad_sub} holds for $k'=k+T_0$ by the same argument as the derivation of  inequality \eqref{eq:LipchitzFuc}. Thus we prove Condition \ref{con:adagrad_sub} for $k'=T_0,T_0+1,\ldots$.

Now that we prove  Condition \ref{con:adagrad_sub} for $k'=0, \ldots,T_0-1, T_0,T_0+1,\ldots$,  we use Lemma \ref{lem: small_w_for_allbsq_sub} and Lemma \ref{lem:log_growth_sub}  to bound $b_k$ denoted by $\bar{b}_{\infty}$:
\begin{align}
\bar{b}_\infty = \eta C\|\vH^{\infty}\|+\frac{4\alpha^2 \sqrt{n}}{\eta {\lambda_{}}C_1}\left( \norm{\vy-\vu(0)} + \frac{ 2\eta^2  \left(C\|\vH^\infty\| \right)^{2}}{ \alpha^2 \sqrt{n}  } \right)
\end{align}
Thus, using the fact that $b_{T_0}/\eta \geq C\|\vH^{\infty}\|$ with $T_0=\frac{\left(\eta C\|\vH^{\infty} \|\right)^2 -b_0^2 }{\alpha^2 \sqrt{ n\varepsilon} }$ and iteratively substituting  $ \norm{\vy-\vu(t)}^2$, $t = T-1,T-2, \cdots, T_0-1 $  in inequality \eqref{eq:LipchitzFuc} gives
\begin{align}
 \norm{\vy-\vu(T)}^2 
  &\leq \Pi_{t=T_0}^{T}\left( 1-\frac{\eta \lambda_0C_1}{2 b_{t}}\right)\norm{\vy-\vu(T_0-1)}^2 \nonumber \\
  &\leq \exp\left( - (T-T_0+1) \frac{\eta \lambda_0C_1}{2\bar{b}_\infty}  \right) \left(  \norm{\vy-\vu(0)} +    \frac{ 2\eta^2  \left(C\|\vH^\infty\| \right)^{2}}{ \alpha^2 \sqrt{n} } 	\right)^2. \nonumber
\end{align}
For the tolerance $\varepsilon$, the maximum step $T$ can be derived by plugging the upper bound $\bar{b}_{\infty}$ into above inequality.

\subsection{Proof of Lemmas}
\label{part3}

Proof of Lemma \ref{lem: small_w_for_allbsq_sub}  and \ref{lem: small_w_for_allbsq1_sub} are given in subection \ref{subsec:ingredients} \\
\textbf{Proof of Lemma \ref{lem:log_growth_sub}}
For $b_{T_0}/\eta \leq C\|\vH^{\infty}\|$, we recalculate the $Term1$ in \eqref{eq:3terms}:
\begin{align*}
Term 1& \leq \left( 1+\frac{ \eta^2}{2b_k^2} \|\mat{H}(k)\|^2 \right)\norm{\vect{y}-\vect{u}(k)}^2 
\end{align*}
After some algebra, we can have
\begin{align}
\norm{\vy-\vu(T_0-1)}  &\leq  \sqrt{ 1+\frac{\eta^2 C{\|\vH^\infty\|^2} }{b^2_{T_0-1}} }\norm{\vy-\vu(T_0-2)} \nonumber\\
&\leq  \norm{\vy-\vu(T_0-2)} +\frac{\eta \sqrt{ C \|\vH^\infty\|^2} }{b_{T_0-1}} \norm{\vy-\vu(T_0-2)} \nonumber\\
 &\leq \norm{\vy-\vu(0)} + \frac{ \eta \sqrt{ C \|\vH^\infty\|^2}}{ \alpha^2 \sqrt{n} } \sum_{t=0}^{T_0-2}  \frac{ \alpha^2 \sqrt{ n} \norm{\vy-\vu(t)} }{  \sqrt{ \alpha^2\sqrt{ n}\sum_{\ell=0}^{t} \| \vy-\vu(\ell)\|+b_0^2}}  \nonumber\\
    &\leq \norm{\vy-\vu(0)} +   \frac{ 2\eta \sqrt{ C \|\vH^\infty\|^2}}{ \alpha^2  \sqrt{n} } \sqrt{ \alpha^2\sqrt{ n}  \sum_{\ell=0}^{T_0-2} \| \vy-\vu(\ell)\|+b_0^2} \nonumber \\
        &\leq \norm{\vy-\vu(0)} +  \frac{ 2\eta^2\left(C\|\vH^\infty\| \right)^{2}}{ \alpha^2  \sqrt{n}}  \nonumber
\end{align}

\section{Proof of Propositions}
\label{sec:proof_prop}
\textbf{Proposition \ref{lem:grad_to_training_sub}} If $\lambda_{min} (\vH)\geq \frac{\lambda_0}{2}$, then $\|\vy -\vu\| \leq \frac{ \sqrt{2m}}{\sqrt{\lambda_0}}\max_{r\in[m]}\| \frac{\partial L (\vW)}{\partial \vw_r}\|.$

\begin{proof}
%\textbf{Proof of Lemma \ref{lem:grad_to_training}} 
For $a_r \sim \text{unif}(\{-1,1\})$ , we have
% $$\frac{1}{\| \mathbf{D}^{-1}\|_1}= \sum_{r=1}^m \frac{1}{\frac{1}{a_r^{2}}}  $$
  \begin{align}
\max_{r\in[m]}\|\frac{\partial L(\vW)}{\partial \vw_r}\|^2
	    & =  \frac{1}{m}\max_{r\in[m]}\norm{
	\sum_{i=1}^n(y_i-u_i)a_r\vx_i\mathbb{I}_{\{\vw_r^\top \vx_i \ge 0\} } }^2 
	 \nonumber\\
&= \frac{1}{m} \max_{r\in[m]}\left( \sum_{i,j}^n (u_i-y_i)(u_j-y_j)\langle{\vx_i,\vx_j\rangle} \mathbb{I}_{\{\vw_r^T\vx_i\geq0,\vw_r^T\vx_j\geq0\}}\right) \nonumber\\
&\geq \frac{1}{m} \left( \sum_{i,j}^n (u_i-y_i)(u_j-y_j)\langle{\vx_i,\vx_j\rangle} \frac{1}{m} \sum_{r=1}^m\mathbb{I}_{\{\vw_r^T\vx_i\geq0,\vw_r^T\vx_j\geq0\}}\right) \nonumber\\
&=  \frac{1}{m}(\vu-\vy)^\top\vH(\vu-\vy) \nonumber\\
&\geq \frac{\lambda_0}{2m}\| \vu-\vy\|^2\nonumber
  \end{align}
where the last inequality use the condition that 
$
    \lambda_{\min}(\vH)\geq\frac{\lambda_0}{2}. 
$
\end{proof}
\textbf{Proposition \ref{prop:adasq}}
Let $L = \eta C \|\vH^{\infty}\|$  and $T_0\geq 1$ be the first index such that $b_{T_0} \geq  L $.  Consider the update: $b_{k+1}^2 = b_{k}^2 +  \alpha^2 \sqrt{n}\|\vy-\vu(k) \| ^2$.
Then for every $r \in [m]$, we have 
\begin{align*}
\|\vw_r(k+1)-\vw_r(0)\|_2
&\leq 
  \frac{\eta\sqrt{2(k+1)}}{\alpha^2 \sqrt{m}}\sqrt{  1+ 2\log \left(\frac{C \eta \|\vH^{\infty}\| }{b_0}\right)}
\end{align*}

\begin{proof}
For the upper bound of $\|\vw_r(k+1)-\vw_r(0)\|_2$ when $b_{t}/\eta < C \|\vH^{\infty}\|, t = 0,1, \cdots, k$ and $k\leq T_0-2$, we have
\begin{align*}
 \sum_{t=0}^{k}  \frac{\norm{\vy-\vu(t)}_2^2}{b_{t+1}^2}  
 &\leq   \frac{1}{\alpha^2 \sqrt{n}  }\sum_{t=0}^{k}  \frac{\alpha^2 \sqrt{n}  \norm{\vy-\vu(t)}_2^2/b_0^2 }{\alpha^2 \sqrt{n} \sum_{\ell=0}^{t} \| \vy-\vu(\ell)\|_2^2/b_0^2 +1} \nonumber \\
&\leq    \frac{1}{\alpha^2 \sqrt{n}  } \left( 1+ \log \left( \alpha^2 \sqrt{n}\sum_{t=0}^{k} \| \vy-\vu(t)\|_2^2/b_0^2 +1\right)\right) \\
&\leq    \frac{1}{\alpha^2 \sqrt{n}  }\left( 1+ 2\log \left(b_{T_0-1}/b_0\right)\right) 
\end{align*}
where the second inequality use Lemma 6 in \cite{ward2018adagrad}.Thus
\begin{align*}
\|\vw_r(k+1)-\vw_r(0)\|_2 
&\leq  \frac{\eta \sqrt{n}}{\sqrt{m}}\sqrt{(k+1) \sum_{t=0}^{k}  \frac{\norm{\vy-\vu(t)}_2^2}{b_{t+1}^2}  } 
\leq   \frac{\eta\sqrt{2(k+1)}}{\alpha^2 \sqrt{m}}\sqrt{  1+ 2\log \left(\frac{C \eta \|\vH^{\infty}\| }{b_0}\right)} . 
\end{align*}
\end{proof}

\section{Technical Lemmas}
\label{sec:basic}
\begin{lem}
\label{lem:increase_sub}
Fix $\varepsilon \in (0,1]$, $L > 0$, $\gamma>0$.   For any non-negative $a_0, a_1, \dots, $ the dynamical system
$$
b_0 > 0; \quad  \quad b_{j+1}^2 = b_j^2 +  \gamma a_j
$$
has the property that after 
{$N = \lceil{ \frac{ L^2-b_0^2}{ \gamma\sqrt{\varepsilon}} \rceil}+1$}
iterations, either $\min_{k=0:N-1}  a_k  \leq  \sqrt{\varepsilon}$, or $b_{N} \geq  L$. 
\end{lem}

\begin{lem}
\label{lem:sqrtsum}
 For any non-negative $a_1,\cdots, a_T$, such that $a_1 > 0$, 
$$ 
\sum_{\ell=1}^T \frac{ a_{\ell} }{ \sqrt{\sum_{i=1}^{\ell} a_i }} \leq  2\sqrt{\sum_{i=1}^{T} a_i}.
$$
\end{lem}
Since the above two lemmas correspond to Lemma 7 and  Lemma 8 in \cite{ward2018adagrad}, we omit their proofs.

\begin{prop} \label{prop:a1}
Under Assumption \ref{asmp:norm1} and Assumption \ref{asmp:lambda_0}, then
$$\max_{r\in[m]}\norm{\frac{\partial L(\vW{})}{\partial \vw_r}}_{2} 
	\le \frac{\sqrt{n}}{\sqrt{m}} \norm{\vy-\vu
	 }_2 .$$
%\begin{align}\label{eq:grad_bound}	
%\end{align}
\end{prop}
The proof is straightforward as follows
	 $$\max_{r\in[m]}\norm{\frac{\partial L(\vW{})}{\partial \vw_r}}
	\le \frac{1}{\sqrt{m}}\sqrt{\sum_{i=1}^{n} |y_i-u_i|^2 }\sqrt{\sum_{i=1}^{n} \|\vx_i\|^2 }  \le \frac{\sqrt{n}}{\sqrt{m}}\norm{\vy-\vu
	 }$$
Observe that at initialization, we have following proposition
\begin{prop}  \label{prop:a2}
Under Assumption \ref{asmp:norm1} and  \ref{asmp:lambda_0}, with probability $1-\delta$ over the random initialization,
\begin{align*}
  \norm{\vect{y}-\vect{u}(0)}^2 \leq \frac{n}{\delta}.
 \end{align*} 
\end{prop}
We get above statement by Markov's Inequality  with following
\begin{align*}
& \expect\left[\norm{\vect{y}-\vect{u}(0)}^2\right]
 =\sum_{i=1}^{n} (y_i^2 + 2y_i \expect \left[f(\mat{W}(0),\vect{a},\vect{x}_i)\right] + \expect\left[f^2(\mat{W}(0),\vect{a},\vect{x}_i)\right]) 
 =\sum_{i=1}^n (y_i^2 + 1) = O(n).
\end{align*}
Finally, we analyze  the upper bound of the maximum eigenvalues of Gram matrix that plays the most crucial role in our analysis.  Observe that
\begin{align*}
\norm{\mat{H}^{\infty}} 
 &=\sup_{\|\vv\|_2=1} \sum_{i,j} v_i v_j\langle{ \vx_i,\vx_j \rangle} \frac{1}{m} \sum_{r=1}^m \indict_{\left\{\vect{w}_r(0)^\top \vect{x}_{i} \ge 0, \vect{w}_r(0)^\top \vect{x}_{j}\ge 0 \right\}} \leq\sqrt{ \sum_{i\neq j}|\langle{ \vx_i,\vx_j \rangle}|^2} +1
\end{align*}
If the data points are pairwise uncorrelated (orthogonal), i.e., $\langle{\vx_i,\vx_j\rangle} = 0, i\neq j$, then  the maximum eigenvalues  is close to 1, i.e., $ \norm{\mat{H}^{\infty}} \leq 1$.  In contrast, we could have  
$\norm{\mat{H}^{\infty}}\le n$ if data points are pairwise highly correlated (parallel), i.e., $\langle{\vx_i,\vx_j\rangle} = 1, i\neq j$.

\begin{table}[H]
\caption{Some notations of parameters to facilitate understanding  the proofs in Appendix B and C}
\label{table}
\centering
\begin{tabular}{ c|c |l }
\hline \toprule
 {Expression}&{Order} &{First Appear} \\ \hline 
 \toprule
 $c$   is a small value, say less than $0.1$ & $ O\left( 1\right)$    & {Lemma \ref{lem:close_to_init_small_perturbation}
} %, the minimum distance for the positiveness of $\vH$
       \\
       \midrule
$R=\frac{c\lambda_0\delta }{n^2}  $& $ O\left( \frac{\lambda_0\delta}{n^2}\right)$        & {Lemma \ref{lem:close_to_init_small_perturbation}
} %, the minimum distance for the positiveness of $\vH$
       \\
       \midrule
$R'= \frac{4\sqrt{n} \norm{\vect{y}-\vect{u}(0)}}{ \sqrt{m} \lambda_0} $& $O\left( \frac{n}{\sqrt{m\delta} \lambda_0}\right) $         &{Lemma \ref{cor:dist_from_init}}
%, the distance of $\norm{\vect{w}_r(k'+1)-\vect{w}_r(0)}$ for gradient descent}     
 \\   
\midrule

 $C_1 = 1-  2 \left( \frac{1}{ \sqrt{n} }+1 \right)\frac{n^2R }{ \lambda_0 \delta}  $& $ O\left(1\right)$       &{Equation \eqref{eq:keyforadagrad}, Condition \ref{con:adagrad_sub}}      
  \\ 
  \midrule
$C=\frac{\left(1+ \frac{2n^2R}{\delta \lambda_0}\right) \left( \|\vH(0)\| + \frac{4n^2R}{\sqrt{2\pi}\delta} \right)+ 2\left(  \frac{1}{n} +1\right) \frac{ ( n^{2} R)^2 }{   \lambda_0\delta^2}}{C_1 \| \mat{H}^{\infty}\|} $& $O\left(1\right)$        &{Equation \eqref{eq:keyforadagrad}, Condition \ref{con:adagrad_sub}}        \\
\midrule
 $\widetilde{R}= \frac{4 \sqrt{n} }{\sqrt{m} \lambda_0 C_1}\norm{\vy-\vu(0)} $& $O\left( \frac{n}{\sqrt{m \delta} \lambda_0}\right) $          &{Lemma \ref{lem: small_w_for_allbsq_sub}}
     \\  
     \midrule
 $\hat{R}= \frac{2\eta^2 C{\|\vH^{\infty}\|}}{ \alpha^2\sqrt{ m}} $& $O\left( \frac{\eta^2\|\vH^{\infty}\| }{\alpha^2\sqrt{m} \lambda_0}\right) $           &{Lemma \ref{lem: small_w_for_allbsq1_sub}}
\\ 
 \midrule
%\hline
\end{tabular}
\end{table}

\end{document}